\documentclass[letterpaper]{article} 
\usepackage{aaai2027}  
\usepackage[hyphens]{url}  
\usepackage{graphicx} 
\usepackage{natbib}  
\usepackage{caption} 
\usepackage{algorithm}
\usepackage{algorithmic}
\usepackage{url}            
\usepackage{amsfonts}       
\usepackage{nicefrac}       
\usepackage{microtype}      
\usepackage{xcolor}         
\usepackage{enumitem} 
\usepackage{amsmath} 
\setcitestyle{numbers,square}
\setlist[itemize]{noitemsep, topsep=0pt}
\usepackage{diagbox} 
\usepackage{float}
\usepackage{tablefootnote}
\usepackage{multirow}
\usepackage{multicol}
\usepackage{adjustbox}
\usepackage{subfig}
\usepackage{longtable}
\usepackage{makecell}
\usepackage{amssymb}
\usepackage{amsthm}      
\usepackage{amssymb} 
\theoremstyle{plain}    
\newtheorem{remark}{Remark}

\newtheorem{proposition}{Proposition}

\newtheorem{corollary}{Corollary}
\theoremstyle{definition}

\newtheorem{assumption}{Assumption}
\usepackage{mathrsfs}
\usepackage{newfloat}
\usepackage{listings}
\DeclareCaptionStyle{ruled}{labelfont=normalfont,labelsep=colon,strut=off} 
\floatstyle{ruled}
\newfloat{listing}{tb}{lst}{}
\floatname{listing}{Listing}

\usepackage{booktabs}

\title{AutoFed: Personalized Federated Traffic Prediction via Adaptive Prompt}
\author{
    Zijian Zhao \textsuperscript{\rm 1},
    Yitong Shang \textsuperscript{\rm 1},
    Sen Li \textsuperscript{\rm 1, \rm 3}\thanks{Corresponding Author: Sen Li}
}
\affiliations{
    \textsuperscript{\rm 1}The Hong Kong University of Science and Technology
    \textsuperscript{\rm 3}The Hong Kong University of Science and Technology (Guangzhou)

}

\begin{document}

\maketitle

\begin{abstract}
Accurate traffic prediction is essential for Intelligent Transportation Systems, including ride-hailing, urban road planning, and vehicle fleet management. However, due to significant privacy concerns, most existing methods rely on local training, resulting in data silos and limited knowledge sharing. Federated Learning (FL) enables privacy-preserving collaborative training, but standard FL struggles with non-independent and identically distributed (non-IID) data across clients. Personalized Federated Learning (PFL) has emerged as a promising paradigm, yet current PFL frameworks require extensive adaptation for traffic prediction tasks, such as graph feature engineering and network architecture design. Moreover, many prior methods depend on dataset-specific hyper-parameter optimization, which impedes practical deployment. To address these challenges, we propose AutoFed, a novel PFL framework for traffic prediction that reduces the dataset-specific manual design required by prior work, most notably hand-crafted graph construction and the manual selection of pattern-extraction filters. Inspired by prompt learning, AutoFed introduces a federated representor that employs a client-aligned adapter to distill local data into a compact, globally shared prompt matrix. This prompt then conditions a personalized predictor, allowing each client to benefit from cross-client knowledge while maintaining local specificity. Because knowledge is shared through this prompt rather than through aggregated parameters, AutoFed naturally supports clients whose road networks differ in the number of nodes, and its per-round communication is independent of the graph size. Extensive experiments on real-world datasets demonstrate that AutoFed consistently achieves competitive performance across diverse scenarios while keeping communication overhead low. The code of this paper is provided at anonymous repository \url{https://anonymous.4open.science/r/AutoFed-3440}.
\end{abstract}


\section{Introduction}

Traffic Prediction (TP) is a cornerstone of modern Intelligent Transportation Systems (ITS), enabling critical applications such as real-time ride-hailing dispatch, dynamic pricing, urban infrastructure planning, and congestion management. However, the development of robust TP models faces a fundamental challenge: although traffic data is abundant, it is also highly sensitive and fragmented \cite{zhang2024survey}. Stringent privacy regulations and commercial interests often result in this data being confined to isolated silos maintained by various municipal authorities (e.g., agencies in different jurisdictions) or private companies (e.g., Uber, Lyft). As a result, centralized training on aggregated datasets is generally infeasible, and most practical solutions must rely on local training with limited data. This siloed approach leads to poor generalization and an inability to capture broader, transferable patterns essential for effective traffic prediction.

Federated Learning (FL) has emerged as a promising paradigm for collaboratively training models without sharing raw data, thereby preserving privacy. However, conventional FL algorithms such as FedAvg \cite{fedavg} are built on the assumption that client data are Independent and Identically Distributed (IID), an assumption that is rarely satisfied in the context of traffic prediction. In reality, traffic patterns exhibit significant heterogeneity (non-IID) across different regions, driven by variations in urban layout, population density, and economic activity. Training a single global model on such non-IID data can lead to client drift and suboptimal performance for all participants. Personalized Federated Learning (PFL) addresses this challenge by allowing clients to learn customized models while still leveraging collective knowledge from the federation \cite{tan2022towards}. As a result, PFL has become a particularly promising framework for achieving realistic, privacy-preserving traffic prediction.

Nevertheless, existing PFL frameworks are not directly suited for the unique challenges of TP. Compared to standard time series prediction tasks, TP faces significant challenges, including complex spatiotemporal dependencies, non-stationary fluctuations, and numerous external features \cite{aouedi2025deep}. This necessitates the adaptation of standard PFL frameworks to meet the specific demands of TP tasks, where traditional methods often fail to incorporate domain-specific needs. Unfortunately, these adaptation processes frequently rely on complex manual efforts, such as intricate graph construction \cite{shang2025security,shang2025II,yuan2022fedstn} or hyper-parameter tuning \cite{hu2024fedgcn,zhou2025fedtps,zhou2024traffic}. For instance, the filter method and pattern amount in \cite{hu2024fedgcn,zhou2025fedtps} can result in performance variations of over 5\% based solely on parameter adjustments mentioned in their papers. However, optimal settings can vary significantly between datasets, and a single configuration can lead to substantial performance variability. The setting process is often contingent upon prior knowledge of the dataset and client configurations or expert insights, which can be challenging to obtain in practice, creating significant barriers to robust and scalable deployment.

A deeper obstacle lies beneath these manual efforts. Most PFL methods share knowledge by aggregating model parameters, yet in traffic prediction different clients own road networks with different numbers of nodes, so any parameter tied to the node axis, such as the adaptive node embedding or adjacency of a graph model, has an incompatible shape across clients and cannot be averaged element-wise. Prior work sidesteps this either by assuming homogeneous clients or by imposing a hand-crafted unified graph, which is precisely what drives the manual design burden above. This motivates a different route: instead of aggregating node-dependent parameters, share a single representation whose size does not depend on the graph.

To this end, we propose AutoFed, a novel PFL framework for traffic prediction. AutoFed pairs a Personalized Predictor (PP) that adapts to local data distributions with a Federated Representor (FR) that distills each client's data into a compact, globally shared prompt matrix; this prompt conditions the PP as a prefix token, so knowledge is shared through the prompt while every node-dependent parameter stays private. We adopt the Adaptive Graph Convolutional Recurrent Network (AGCRN) \cite{bai2020adaptive} as an off-the-shelf backbone rather than as a contribution, in order to isolate our novelty in the federated collaboration mechanism itself. Concretely, our contributions are threefold. \emph{(i)} We introduce a federated representor that aligns heterogeneous local representations into a shared prompt space via a client-aligned adapter, enabling clients with differing graph sizes to collaborate in a common parameter space without any hand-crafted unified graph, a setting where standard parameter-aggregation PFL does not directly apply.
\emph{(ii)} By generating the prompt with a learnable denoiser and adapter, AutoFed replaces the hand-crafted graph construction and pattern-filter selection that prior work must re-tune per dataset with a data-driven module, removing this dataset-specific design while keeping only a few standard architectural and optimization hyper-parameters that are fixed across all datasets. \emph{(iii)} Extensive experiments on real-world travel demand and traffic flow prediction tasks show that AutoFed consistently achieves competitive accuracy across diverse federated scenarios while maintaining low communication overhead, and a representation-level analysis further reveals how the learned prompt aligns clients and what traffic information it encodes.

\section{Related Work}

\subsection{Traffic Prediction}
Serving as a key component of ITS, TP encompasses various tasks such as travel demand prediction, traffic flow prediction, and travel time prediction. Early research treated each traffic node as an independent entity, framing TP as a general time series prediction task. Many methods were developed based on conventional single-dimensional time series approaches, including statistical methods \cite{liu2011discovering}, cluster-based pattern matching \cite{zhang2013improved}, and neural networks \cite{yang2019traffic}. However, these approaches often overlook the spatial relationships among traffic nodes. For instance, when a traffic jam occurs in one area, neighboring regions are likely to experience increased traffic flow as well. Subsequently, most approaches adopted a paradigm that first extracts spatial features within each time frame and then employs time series networks for prediction. For example, \cite{liu2011discovering, asif2013spatiotemporal,ke2017short} proposed a CNN-LSTM network for traffic flow prediction. However, CNN-based spatial extraction struggles to adequately represent the non-Euclidean nature of traffic networks, resulting in challenges in accurately learning robust spatial features \cite{zhou2024traffic, zhou2025fedtps}. Recently, Graph Neural Networks (GNNs) have emerged as a promising solution, effectively capturing the relationships among nodes in traffic networks. An example is the AGCRN \cite{bai2020adaptive}, which uses a GCN-LSTM structure and employs adaptive adjacency matrices to capture spatial dependencies, allowing for end-to-end learning. Additionally, STWave \cite{fang2023spatio} introduced a novel decoupling approach to separate traffic data into stationary and non-stationary components, utilizing both temporal and spatial attention methods to capture relationships accurately.

Despite these advancements, most of these methods rely on centralized training, neglecting the privacy considerations inherent in traffic data, which are increasingly recognized by data privacy regulations such as the General Data Protection Regulation (GDPR) and the California Consumer Privacy Act (CCPA). Recently, some research has begun to focus on FL-based paradigms to address privacy concerns. For instance, Shang et al. \cite{shang2025security,shang2025II} proposed both horizontal and vertical methods based on FedAvg \cite{fedavg}, in which a carefully designed traffic graph efficiently captures the relationships among traffic nodes. Additionally, Hu et al. \cite{hu2024fedgcn} introduced FedGCN, which considers the relationship between graph nodes and external factors through a novel MendGCN module. In the realm of PFL methods, Zhou et al. \cite{zhou2024traffic,zhou2025fedtps} developed FedTPS, which utilizes a traffic pattern repository to share knowledge among clients. \emph{However, most of these methods depend on meticulous manual feature design and hyper-parameter optimization tailored to specific clients or datasets, which hinders their efficient application and adaptation to distinct contexts.}

\subsection{Federated Learning}
In recent years, FL has been widely applied in high-privacy and resource-limited scenarios \cite{cai2025feedsign}, including medical recognition \cite{pfitzner2021federated}, traffic prediction \cite{shang2025explainable}, and autonomous driving \cite{kou2025fast}. However, standard FL methods like FedAvg rely on the assumption of IID data among clients, which is not always feasible in practice. According to \cite{pei2024review}, solutions for non-IID data distributions can be categorized into three types: data-based, framework-based, and model-based methods. Among these, PFL, as a model-based approach, shows promising performance. In PFL, clients share only portions of the model to leverage common knowledge while retaining personalized parameters or structures to adapt to local data distributions.

Starting with FedPer \cite{fedper}, different clients share upstream layers for robust feature extraction while keeping personalized downstream layers for final output, which is also a well-recognized paradigm in the theory of transfer learning \cite{zhuang2020comprehensive}. Following this, various methods have introduced innovative technologies such as clustering \cite{chang2025dual}, where clients within a single cluster share data distribution similarities; meta-learning \cite{fallah2020personalized}, which aims to find a suitable starting point for all clients; and hyper-networks \cite{scott2024pefll}, which directly output personalized network parameters with a high capacity for expansion to new clients. Moreover, PFL can be effectively utilized in scenarios where different clients have varying model structures, as it allows for the aggregation of common sub-networks. This flexibility is particularly advantageous when clients possess different computational resources for varying model scales. \emph{As mentioned in the introduction, due to the unique challenges in TP, it is essential to adapt these methods for specific tasks. However, this adaptation still encounters the challenges faced by conventional TP models discussed in the previous subsection.}

\subsection{Prompt Learning}
Recently, the success of Large Language Models (LLMs) \cite{chen2026overview} has led to the effective application and adaptation of various related technologies, such as Masked Language Model (MLM) pre-training \cite{zhao2025csi}, Retrieval-Augmented Generation (RAG) \cite{hanretrieval}, and LLMs themselves \cite{jin2024time}, in time series analysis. Among these technologies, prompt learning is recognized for its efficiency and ease of implementation. In the context of LLMs, prompts play a crucial role in directly influencing the subsequent output decoding, a concept well-established since the development of T5 \cite{raffel2020exploring}. Subsequently, prompt tuning has emerged as a promising solution for enhancing or fine-tuning models with low cost by utilizing learnable prompt tokens \cite{li2025survey}.

In the field of time series analysis, most research has concentrated on designing prompts to adapt pre-trained LLMs or their architectures for specific tasks \cite{caotempo,xue2023promptcast}. For example, \cite{huang2025timedp,zhou2024traffic,zhou2025fedtps} employ learnable prototypes (or patterns) as guiding prompts for the decoding process. However, the number of prototypes can significantly impact model performance, and the optimal quantity varies across datasets \cite{zhou2024traffic,zhou2025fedtps}. \emph{In contrast, this paper proposes a neural network-based prompt generator, eliminating the need for prototypes.} Furthermore, while some graph-based works have also introduced prompt learning \cite{li2024graph}, their focus is primarily on designing spatial prompt representations, which differs from our emphasis on the temporal dimension.

\section{Methodology} \label{sec:methodology}

\begin{figure}[htbp]
\centering 
\includegraphics[width=0.5\textwidth]{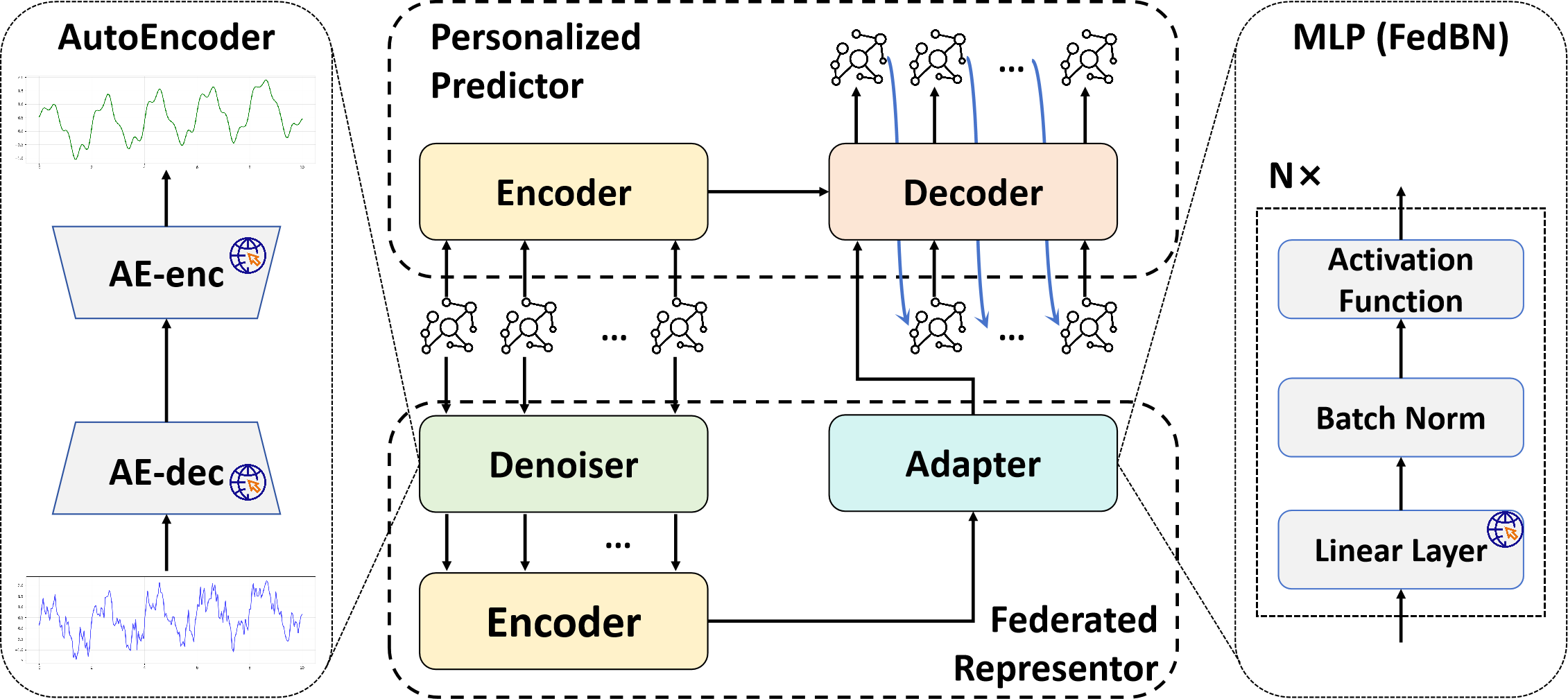}
\caption{Network Architecture: The network consists of the PP and FR. PP utilizes a graph time series network with an encoder-decoder structure. FR employs an AE-based denoiser for robust feature extraction, a graph time series encoder for feature compression, and an client-aligned adapter for transferring local representations to global representations, providing a guided prompt matrix for the decoder in PP. In the figure, only the modules with the ``earth" icon are shared among clients.}
\label{fig:main}
\end{figure}

\subsection{Problem Setup}

In this paper, we study the federated TP task, which involves $m$ clients $\mathcal{C} = \{C_1,C_2, \ldots, C_\mathcal{N} \}$ with non-IID data and a centralized server $S$. For each client $i\in\{1,\ldots,\mathcal{N}\}$, the transportation network can be represented as $G_i = (V_i, E_i)$, where $V_i$ and $E_i$ denote the nodes and edges, respectively. Additionally, $A_i \in \mathbb{R}^{|V_i|,|V_i|}$ is the weighted adjacency matrix, representing the relationship (e.g. spatial, similarity, and dependencies) among traffic nodes, where  and $|\cdot|$ represents the size of set. Specifically, each client possesses a private dataset $D_i = \{X_i, Y_i\}$, where $X_i \in \mathbb{R}^{N_i, \mathcal{T}, |V_i|, K}$ represents the historical traffic features, and $Y_i \in \mathbb{R}^{N_i, \mathscr{T}, |V_i|}$ is the ground truth for future demand. Specifically, $N_i$ denotes the sample size for client $i$, $\mathcal{T}$ and $\mathscr{T}$ are the lengths of the historical information sequence and the future horizon to be predicted, respectively, and $K$ is the feature dimension.

Specifically, we consider a PFL scenario, where the network of each client $i$ at round $m$ can be expressed as $\text{f}(\cdot; \theta_i^m, \Theta^m)$, where $\theta_i^m$ represents the private part of the network parameters of client $i$, which learns the personalized aspects of the data in their own data distribution, while $\Theta^m$ denotes the shared part of the network parameters, which learns a global general representation for all clients. Consequently, the target of PFL is defined as:
\begin{equation}
\begin{aligned}
\arg \min_{\theta_1, \theta_2, \ldots, \theta_\mathcal{N}, \Theta} \sum_{i=1}^\mathcal{N} \frac{|V_i|}{\sum_{i=1}^\mathcal{N}|V_i|} \text{L}(\text{f}(X_i; \theta_i, \Theta), Y_i) \ ,
\label{eq:target}
\end{aligned}
\end{equation}
where $\text{L}(\cdot, \cdot)$ is the loss function.

\subsection{Method Overview}
The network structure is depicted in Fig. \ref{fig:main}, comprising two main components: the Personalized Predictor (PP) and the Federated Representor (FR). The PP utilizes an encoder-decoder-based graph time series network that is trained locally, allowing it to adapt to local data distributions. During inference, the encoder receives historical data $x \in \mathbb{R}^{\mathcal{T}, n, K}$ as input, while the decoder predicts future traffic data $\hat{y} \in \mathbb{R}^{ \mathscr{T}, n}$ in an Auto-Regressive (AR) manner. For client $i$, the input $x$ is derived from its dataset $X_i$, where $n$ represents the number of nodes ($|V_i|$). Specifically, the FR generates a prompt matrix $p_g \in \mathbb{R}^{n,k}$ based on the input $x$ (where $k$ is the hidden dimension), which serves as a prefix token for the decoder. This design is motivated by the intuition that different clients may share similar traffic patterns \cite{zhou2024traffic,zhou2025fedtps}, allowing the prefix prompt to convey pertinent pattern information from the input $x$ and guide the decoding process. During training, the parameters of the FR are partially shared among clients, facilitating the learning of a global common traffic pattern representation. Simultaneously, personalized components are also employed to adapt to local data distributions like PP.

\subsection{Personalized Predictor}

Although our framework is ultimately dependent on a graph time series model, we recommend utilizing methods with adaptive graph structures, such as the AGCRN \cite{bai2020adaptive}, which is also employed in the experiments of this paper. This recommendation is based on the following considerations. Currently, most traffic prediction models rely on manually designed graphs \cite{shang2025security,shang2025explainable}. While these approaches yield promising results, constructing such graphs is time-consuming and requires careful consideration of multiple factors, including spatial connections, distance relationships, and node similarities. Furthermore, the effectiveness of each graph feature may vary across different regions or datasets. In contrast, AGCRN adapts the graph structure through a learnable parameter, $E \in \mathbb{R}^{n,k}$.
The AGCRN process can be represented as:
\begin{equation}
\begin{aligned}
\tilde{A} & = \text{Softmax}(\text{ReLU}(EE^T)) \ , \\
u_t &= \sigma(\operatorname{GCN}_{u}(x_t, H_{t-1}, \tilde{A}))\ , \\
r_t &= \sigma(\operatorname{GCN}_{r}(x_t, H_{t-1}, \tilde{A}))\ , \\
C_t &= \tanh(\operatorname{GCN}_{C}(x_t, (r_t \odot H_{t-1}), \tilde{A}))\ , \\
H_t &= u_t \odot H_{t-1} + (1 - u_t) \odot C_t\ ,
\label{eq:GRU}
\end{aligned}
\end{equation}
where the matrix $\tilde{A} \in \mathbb{R}^{n,n}$ is the adaptive adjacency matrix, $x_t \in \mathbb{R}^{n,K}$ represents the input $x$ at frame $t$ (i.e. $x=\{x_1,x_2, \ldots,x_{\mathcal{T}}\}$), and $u_t, r_t, C_t, H_{t-1} \in \mathbb{R}^{n,k}$ represent the update gate, reset gate, candidate state, and previous hidden state of the GRU at frame $t$, respectively. Specifically, we employ the same AGCRN structure both the encoder and decoder in the PP.

\subsection{Federated Representor}
As shown in Fig. \ref{fig:main}, the FR consists of three components: an Auto-Encoder (AE) denoiser for robust feature extraction, a graph time series encoder for further information extraction and compression, and a client-aligned adapter to transfer the local feature representation to a global common feature prompt, which serves as guidance for the PP.

\begin{itemize}[left=0pt]
\item \textbf{Auto-Encoder Denoiser:} 
The goal of the FR is to generate a guiding prompt that captures the traffic pattern information, which necessitates a focus on stable features while ignoring random noise. While some previous works have utilized low-pass filters \cite{zhou2024traffic,zhou2025fedtps,fang2023spatio}, selecting the appropriate filtering method and its hyper-parameters can be challenging, directly affecting model performance \cite{zhou2024traffic,zhou2025fedtps}. Additionally, these methods rely on the assumption that low-frequency components contain more stable patterns, whereas high-frequency components contain more noise. However, the specific threshold between these components is often unclear. To address these issues, we propose using an AE-based denoiser that can learn stable patterns autonomously. This approach has proven effective in various fields, including signal processing \cite{nguyen2025robust} and image denoising \cite{lee2021noise}. According to most of the previous traffic time series research \cite{boto2010wavelet,zhou2025fedtps,qin2024traffic}, the traffic data $x$ can be viewed as:
\begin{equation}
\begin{aligned}
x = \tilde{x} + \xi \ ,
\label{eq:noise}
\end{aligned}
\end{equation}
where $\tilde{x}$ represents regular traffic patterns, and $\xi$ represents stochastic noise, with the assumption that $\mathbf{E}[\xi] = 0$. Then the AE-based denoising process can be represented as:
\begin{equation}
\begin{aligned}
p & = \text{AE-enc}(x) \ , \\
\hat{x} & = \text{AE-dec}(p) \ ,
\label{eq:ae}
\end{aligned}
\end{equation}
where the AE encoder compresses the original input sequence $x \in \mathbb{R}^{\mathcal{T}, n, K}$ into a feature space $p \in \mathbb{R}^{\mathcal{T}, n, k}$, and the AE decoder attempts to recover it back to the original input $\hat{x} \in \mathbb{R}^{\mathcal{T}, n, K}$. The objective of the AE denoiser can be represented as:
\begin{equation}
\begin{aligned}
\delta^* & = \arg \min_\delta \|\hat{x} - x\|_1 \ ,
\label{eq:object}
\end{aligned}
\end{equation}
where $\delta$ denotes the parameters of the AE, and we utilize minimizing the L1 norm $\|\cdot\|_1$ as the target. Since time series data predominantly consists of regular patterns $\tilde{x}$ and irregular noise $\xi$, recovering the noise element can be challenging, allowing the AE to effectively perform denoising. Since denoising is similar across different nodes, we choose to share this module among clients. 
While it is not guaranteed that an AE perfectly separates signal from noise, prior studies in time series \cite{nolan2022multi,kieu2022robust} have shown that under an information bottleneck, AEs tend to prioritize reconstructing low-frequency, high-amplitude components, which often correspond to the underlying signal. Therefore, we employ the AE as a trainable alternative to hand-crafted filters, allowing the model to learn a denoising strategy adapted to the data. The reconstruction loss implicitly encourages the hidden representation $p$ to retain stable patterns while suppressing irregular fluctuations.

\item \textbf{Graph Time Series Encoder:} 
Even though the AE denoiser provides robust stable features, it can be difficult to use the long-sequence $p$ as a prompt for the PP. Therefore, we propose utilizing a graph time series encoder to compress $p$ while extracting additional information. Specifically, we adopt the same encoder structure as the PP, compressing the feature $p$ into a single matrix $p_l \in \mathbb{R}^{n, k}$, referred to as the local feature. Given the differing graph structures and data distributions among clients, we opt for a personalized encoder for each client. 

\item \textbf{Client-Aligned Adapter:} 
To facilitate knowledge sharing among clients, we introduce a client-aligned adapter that transfers the local traffic pattern representation $p_l$ to a global representation $p_g \in \mathbb{R}^{n, k}$. For this process, we utilize a shared MLP. Specifically, we employ the FedBN \cite{li2021fedbn} strategy to tackle the challenges posed by non-IID data, which persists in the local feature space $p_l$. In detail, the parameters of the linear layers are shared among clients, while the batch normalization layers remain distinct, retaining their statistical information, such as the average values and standard deviations, along with the rescaling parameters. 

To summarize, the full process of the PP can be expressed as follows:
\begin{equation}
\begin{aligned}
H_{\mathcal{T}} & = \text{Encoder}(x; \mathbf{0}) \ , \\ 
z & = \text{Decoder}(p_g; H_{\mathcal{T}}) \ , \\
\hat{y} & = \text{MLP}(z) \ ,
\label{eq:decode}
\end{aligned}
\end{equation}
where $\mathbf{0}$ represents a zero matrix that serves as the initial hidden state for the encoder, $p_g$ is the first input token for the decoder, $H_{\mathcal{T}}$ is the final hidden state of the encoder, which serves as the initial hidden state for the decoder, the variable $z \in \mathbb{R}^{\mathscr{T}, n, k}$ is the output of the decoder, and an additional MLP $(\mathbb{R}^{k} \rightarrow \mathbb{R})$ maps $z$ to the final prediction result $\hat{y} \in \mathbb{R}^{\mathscr{T}, n}$. It is important to note that the intermediate output $z$ is necessary because the output dimension of each token in the decoder must match that of the first prompt token $p_g$, due to its AR nature. Note that although the encoder and decoder have the same structure, the inputs ($x \in \mathcal{R}^{\mathcal{T},n,K}$ and $p_g \in \mathcal{R}^{n,k}$) have different dimensions. This is because the decoder follows the AR paradigm, while the encoder processes the full input sequence.

\end{itemize}

\subsection{Training Process}

In considering the training process, we first define the loss function, which consists of two parts: the regression loss of the AE-denoiser and the regression loss of the PP. In this paper, we choose the Mean Absolute Error (MAE) as the loss function. For each client $i$, the two loss functions can be represented as:
\begin{equation}
\begin{aligned}
L^{ae}_i & = \mathbf{E}_{x \in X_i} ||x - \hat{x}||_1 \ , \\
L^{pre}_i & = \mathbf{E}_{y \in Y_i} ||y - \hat{y}||_1 \ ,
\label{eq:mae}
\end{aligned}
\end{equation}
where $||\cdot||_1$ denotes the $L_1$ norm. The overall loss function for client $i$ is then defined as:
\begin{equation}
\begin{aligned}
L_i & = L^{pre}_i + \alpha L^{ae}_i \ ,  \\
\alpha & = \frac{L^{ae}_i}{L^{pre}_i} \ ,
\label{eq:loss}
\end{aligned}
\end{equation}
where $\alpha$ is an adaptive hyper-parameter, inspired by \cite{zhao2025let,chen2018gradnorm,kendall2018multi}. This approach allows the training process for denoising and prediction to adaptively balance: when the denoiser has a high error, the prompt matrix may not be effective, making a large weight for the prediction loss less meaningful since the prompt matrix will change thereafter. Thus, a higher weight for the denoising loss is warranted. This relationship is vice versa as well. The detailed training process follows the conventional PFL paradigm, as shown in Appendix \ref{sec:alg}. We analyze the convergence and stability of this procedure, and interpret the prompt mechanism, in Appendix \ref{sec:theory}; in particular, Proposition \ref{prop:converge} shows that the training converges to a stationary point of Eq.~\eqref{eq:target} and cannot diverge for a suitable step size, and Corollary \ref{cor:fedavg} shows that the attainable objective is never worse than a fully shared FedAvg model.

\section{Experiment}

\subsection{Experiment Setup} \label{sec:setup}

In our experiments, we compare the performance of our method against several benchmarks, including general FL methods (FedAvg \cite{fedavg}, FedProx \cite{fedprox}), PFL methods (FedPer \cite{fedper}, pFedMe \cite{pfedme}), and State-Of-The-Art FL methods focused on traffic prediction (FedTPS \cite{zhou2024traffic,zhou2025fedtps}, FedGCN \cite{hu2024fedgcn}). To ensure fairness, we employ the same AGCRN architecture as the backbone for all methods. The detailed introduction of these methods is shown in Appendix \ref{sec: Detailed Description of Comparative Methods}. Additionally, we utilize local training (where each client trains a personalized model using their own data) as a baseline. We also conduct an ablation study to illustrate the efficacy of each module, evaluating our method when the AE-denoiser and FedBN modules are removed individually. The experiments are conducted on a workstation running Windows 11, equipped with an Intel(R) Core(TM) i7-14700KF processor and an NVIDIA RTX 4080 graphics card.

To validate our method, we conduct two case studies: one for the Travel Demand Prediction (TDP) task and another for the Traffic Flow Prediction (TFP) task, both utilizing real-world datasets. For TDP, we use ride-hailing data from the New York City Taxi and Limousine Commission dataset \cite{deri2015taxi}, which includes demand data for Uber and Lyft across five governmental districts in New York City (NYC). Initially, we consider the entire Uber and Lyft data as two separate clients, leading to scenario (i) S0. However, our experiments reveal that this approach is not optimal. In the current configuration, the graph encompasses the entirety of New York City, which is excessively large and may not accurately reflect real-world operational patterns. In practice, traffic demand in Manhattan has limited impact on Queens, as most ride-hailing activities are confined within individual regions. Combining all districts into a single graph can introduce mutual interference, complicating the training process—particularly for models with adaptive adjacency matrices such as AGCRN. We find that dividing the overall graph into several regional subgraphs actually yields better prediction performance (detailed experimental results are presented in the next subsection). Therefore, we propose treating each company within each district as a separate client, although this decision is motivated by modeling effectiveness rather than privacy concerns. We consider three scenarios: (ii) S1: the five Uber clients are trained together; (iii) S2: the five Lyft clients are trained together; (iv) S3: all Uber and Lyft clients are trained together. For TFP, we employ three datasets from the California Transportation Agencies (CalTrans) Performance Measurement System (PEMS) \cite{pems}, with each dataset (PEMS03, PEMS04, PEMS08) originating from different regions and time periods on California's highways. We divide each dataset into client numbers of 2, 4, 6, 8, and 10 and report the average performance across these setups. 


In the experiments, we split the data into training, validation, and test sets with a ratio of 6:2:2. The batch size is set to 128, and the learning rate is $10^{-3}$ with the Adam optimizer. For federated learning configurations, we set the number of communication rounds to 50 and use one local epoch. All experiments were repeated three times, and the standard deviations are reported in the tables. During experiment, we consider the following evaluation metrics: Mean Absolute Error (MAE), Root Mean Square Error (RMSE), Mean Square Error (MSE), and Mean Absolute Percentage Error (MAPE).

\subsection{Experiment Results}

\begin{table*}[t!]
\centering
\caption{Model Performance on the TDP Task: The best results are highlighted in \textbf{bold}. The value after $\pm$ indicates the standard deviation over three independent runs. This notation is used consistently throughout the subsequent tables.}
\begin{adjustbox}{width=\textwidth}
\begin{tabular}{llcccccccc}
\toprule
\multirow{2}{*}{Type} & \multirow{2}{*}{Method}   & \multicolumn{4}{c}{Uber (S0)} & \multicolumn{4}{c}{Lyft (S0)}  \\
\cmidrule(lr){3-6} \cmidrule(lr){7-10}
& & MAE & RMSE & MSE  & MAPE/\% & MAE & RMSE & MSE  & MAPE/\% \\
\midrule
\multicolumn{2}{c}{Local Training}  & 3.70±0.13 & 6.25±0.27 & 39.17±3.41 & 40.15±0.69 &  2.08±0.10 & 3.31±0.17 & 10.95±1.09 & 51.47±1.00 \\
\midrule
\multirow{2}{*}{FL} & FedAvg \cite{fedavg}    & 3.58±0.09 & 6.08±0.21 & 37.17±2.69 & 39.15±0.35  &   1.96±0.08 & 3.13±0.14 & 9.79±0.89 & 49.41±0.50 \\
& FedProx \cite{fedprox}   & 3.55±0.02 & 5.96±0.06 & 35.68±0.78 & 40.34±0.41 &  1.88±0.08 & 2.97±0.15 & 8.83±0.91 & 49.67±0.46 \\
\midrule
\multirow{2}{*}{PFL} & FedPer \cite{fedper}   & 3.67±0.14 & 6.23±0.31 & 39.04±3.83 & 40.24±1.29 & 1.90±0.05 & 2.99±0.10 & 8.93±0.61 & 50.84±0.15 \\
& pFedMe \cite{pfedme}   & 3.63±0.04 & 6.19±0.13 & 38.42±1.63 & 40.13±0.21 & 2.08±0.11 & 3.32±0.20 & 11.09±1.32 & 51.85±1.11 \\
\midrule
\multirow{2}{*}{TrafficFL} & FedTPS \cite{zhou2024traffic,zhou2025fedtps}   & 3.65±0.15 & 6.23±0.33 & 39.00±4.13 & 39.43±0.81 & 1.95±0.07 & 3.10±0.13 & 9.63±0.79 & 50.58±0.29 \\
& FedGCN \cite{hu2024fedgcn}  & \textbf{3.44±0.01} & \textbf{5.72±0.06} & \textbf{32.83±0.74} & 38.61±0.03 & 1.85±0.03 & 2.94±0.06 & 8.68±0.36 & \textbf{48.57±0.16}  \\
\midrule
\multirow{3}{*}{Ours} & AutoFed    & 3.48±0.11 & 5.92±0.24 & 35.18±2.82 & \textbf{38.44±0.45}  & \textbf{1.84±0.03} & \textbf{2.90±0.05} & \textbf{8.43±0.28} & 48.85±0.27 \\
& w/o AE    & 3.67±0.23 & 6.20±0.38 & 38.62±4.73 & 39.88±0.75 & 2.07±0.17 & 3.30±0.29 & 10.95±1.94 & 50.82±1.80  \\
& w/o FedBN   & 3.66±0.12 & 6.20±0.27 & 38.56±3.29 & 40.20±0.55 & 1.98±0.02 & 3.17±0.04 & 10.04±0.24 & 50.21±0.41  \\
\bottomrule \toprule
\multirow{2}{*}{Type} & \multirow{2}{*}{Method}   & \multicolumn{4}{c}{Uber (S1)} & \multicolumn{4}{c}{Lyft (S2)}  \\
\cmidrule(lr){3-6} \cmidrule(lr){7-10}
& & MAE & RMSE & MSE  & MAPE/\% & MAE & RMSE & MSE  & MAPE/\% \\
\midrule
\multicolumn{2}{c}{Local Training}  & 3.45±0.05 & 5.43±0.10 & 33.69±1.56 & 39.45±0.13 &  2.00±0.04 & 3.06±0.06 & 9.61±0.39 & 49.16±0.19 \\
\midrule
\multirow{2}{*}{FL} & FedAvg \cite{fedavg}  & 3.72±0.16 & 5.96±0.27 & 41.30±4.20 & 40.17±0.10 &  2.02±0.03 & 3.09±0.06 & 9.78±0.38 & 49.85±0.02 \\
& FedProx \cite{fedprox} & 3.63±0.12 & 5.83±0.22 & 39.02±2.40 & 40.09±0.56 & 1.96±0.07 & 3.00±0.12 & 9.54±0.84 & 51.92±0.46 \\
\midrule
\multirow{2}{*}{PFL} & FedPer \cite{fedper} & 3.75±0.15 & 5.95±0.26 & 40.40±3.50 & 40.87±0.37 & 1.95±0.06 & 3.01±0.10 & 9.57±0.67 & 51.03±0.23 \\
& pFedMe \cite{pfedme} & 3.45±0.07 & 5.47±0.14 & 33.94±1.89 & 39.70±0.38 & 1.92±0.05 & 2.95±0.10 & 9.21±0.67 & 50.64±0.44 \\
\midrule
\multirow{2}{*}{TrafficFL} & FedTPS \cite{zhou2024traffic,zhou2025fedtps}  &3.52±0.06 & 5.54±0.09 & 34.75±0.91 & 39.76±0.51  &  1.87±0.02 & 2.86±0.03 & 8.62±0.15 & 50.56±0.30 \\
& FedGCN \cite{hu2024fedgcn} & 3.58±0.11 & 5.71±0.21 & 36.34±2.43 & 39.52±0.39 & 2.02±0.05 & 3.10±0.09 & 9.91±0.56 & \textbf{48.90±0.21} \\
\midrule
\multirow{3}{*}{Ours} & AutoFed  &  \textbf{3.40±0.04} & \textbf{5.34±0.07} & \textbf{32.43±0.92} & 38.89±0.36   & \textbf{1.84±0.03} & \textbf{2.80±0.05} & \textbf{8.37±0.31} & 49.70±0.36 \\
& w/o AE   & 3.46±0.06 & 5.44±0.12 & 33.64±1.44 & \textbf{38.81±0.52} &  1.89±0.03 & 2.91±0.06 & 8.98±0.33 & 49.88±0.34 \\
& w/o FedBN  &  3.43±0.07 & 5.41±0.13 & 32.98±1.25 & 38.75±0.26  &  1.89±0.03 & 2.89±0.06 & 8.84±0.39 & 50.35±0.12 \\
\bottomrule \toprule
\multirow{2}{*}{Type} & \multirow{2}{*}{Method}   & \multicolumn{4}{c}{Uber (S3)} & \multicolumn{4}{c}{Lyft (S3)}  \\
\cmidrule(lr){3-6} \cmidrule(lr){7-10}
& & MAE & RMSE & MSE  & MAPE/\% & MAE & RMSE & MSE  & MAPE/\% \\
\midrule
\multicolumn{2}{c}{Local Training}  & 3.45±0.05 & 5.43±0.10 & 33.69±1.56 & 39.45±0.13 &  2.00±0.04 & 3.06±0.06 & 9.61±0.39 & 49.16±0.19 \\
\midrule
\multirow{2}{*}{FL} & FedAvg \cite{fedavg}   & 3.73±0.12 & 6.02±0.12 & 42.20±0.77 & 41.43±1.45 & 1.93±0.06 & 2.99±0.11 & 9.42±0.69 & 50.09±0.57 \\
& FedProx \cite{fedprox}  & 3.73±0.16 & 6.01±0.31 & 43.09±5.37 & 40.97±1.04 & 2.03±0.07 & 3.12±0.13 & 10.05±0.78 & \textbf{48.81±0.51} \\
\midrule
\multirow{2}{*}{PFL} & FedPer \cite{fedper}  & 3.90±0.20 & 6.17±0.33 & 43.86±4.81 & 41.34±0.63 & 1.97±0.05 & 3.05±0.10 & 9.90±0.67 & 51.55±0.26 \\
& pFedMe \cite{pfedme}  & 3.44±0.03 & 5.42±0.04 & 33.30±0.25 & 39.99±0.13  & 1.90±0.04 & 2.91±0.07 & 8.92±0.41 & 50.69±0.18 \\
\midrule
\multirow{2}{*}{TrafficFL} & FedTPS \cite{zhou2024traffic,zhou2025fedtps}  & 3.45±0.03 & 5.45±0.03 & 33.85±0.20 & 39.24±0.20  & 1.91±0.05 & 2.94±0.09 & 9.13±0.54 & 50.36±0.25 \\
& FedGCN \cite{hu2024fedgcn}  & 3.57±0.07 & 5.66±0.10 & 35.79±0.93 & 40.01±0.62 &  1.95±0.04 & 3.01±0.06 & 9.51±0.39 & 51.32±1.0 \\
\midrule
\multirow{3}{*}{Ours} & AutoFed   & \textbf{3.36±0.04} & \textbf{5.27±0.07} & \textbf{31.56±0.89} & \textbf{38.82±0.49} & \textbf{1.85±0.02} & \textbf{2.82±0.04} & \textbf{8.40±0.19} & 50.33±0.29 \\
& w/o AE   & 3.47±0.06 & 5.48±0.11 & 34.16±1.22 & 39.00±0.19 &  1.87±0.03 & 2.87±0.06 & 8.82±0.37 & 49.59±0.17 \\
& w/o FedBN   & 3.43±0.06 & 5.39±0.08 & 32.93±0.83 & 39.69±0.32  &  1.88±0.03 & 2.88±0.06 & 8.79±0.40 & 50.53±0.18 \\
\bottomrule
\end{tabular}
\end{adjustbox}
\label{tab:exp}
\end{table*}

\begin{table*}[t!]
\centering
\caption{Model Performance on TFP Task for Three Datasets}
\begin{adjustbox}{width=0.75\textwidth}
\begin{tabular}{lllccccc}
\toprule
Dataset & Type & Method & MAE & RMSE & MSE & MAPE/\% \\
\midrule
\multirow{9}{*}{PEMS03} & \multicolumn{2}{c}{Local Training} & 18.01±0.26 & 27.66±0.39 & 907.39±16.28 & 18.11±1.16 \\
\cmidrule(lr){2-7}
 & \multirow{2}{*}{FL} & FedAvg \cite{fedavg} & 17.89±0.24 & 27.70±0.42 & 901.38±20.13 & 18.15±0.88 \\
 &                    & FedProx \cite{fedprox} & 17.81±0.35 & 27.50±0.39 & 885.76±29.52 & 18.36±0.93 \\
\cmidrule(lr){2-7}
 & \multirow{2}{*}{PFL} & FedPer \cite{fedper} & 18.00±0.18 & 27.80±0.56 & 909.52±12.73 & 18.90±0.93 \\
 &                    & pFedMe \cite{pfedme} & 16.99±0.28 & 26.59±0.34 & 853.63±25.99 & 16.98±0.91 \\
\cmidrule(lr){2-7}
 & \multirow{2}{*}{TrafficFL} & FedTPS \cite{zhou2024traffic,zhou2025fedtps} & 17.19±0.16 & 26.89±0.61 & 869.70±12.07 & 17.19±0.40 \\
 &                           & FedGCN \cite{hu2024fedgcn} & 18.82±0.22 & 28.87±0.48 & 985.89±22.28 & 20.35±1.42 \\
\cmidrule(lr){2-7}
 & \multirow{3}{*}{Ours} & AutoFed   & \textbf{16.15±0.17} & \textbf{25.57±0.39} & \textbf{825.39±46.40} & \textbf{16.06±0.48} \\
 &                      & w/o AE    & 16.27±0.12 & 25.58±0.30 & 795.82±11.35 & 16.60±0.44 \\
 &                      & w/o FedBN & 16.93±0.31 & 26.50±0.52 & 849.11±28.88 & 17.27±0.63 \\
\midrule \midrule
\multirow{9}{*}{PEMS04} & \multicolumn{2}{c}{Local Training} & 22.43±0.23 & 33.84±0.31 & 1204.04±8.75 & 16.11±0.39 \\
\cmidrule(lr){2-7}
 & \multirow{2}{*}{FL} & FedAvg \cite{fedavg} & 23.06±0.50 & 34.64±0.39 & 1266.97±41.00 & 16.90±0.87 \\
 &                    & FedProx \cite{fedprox} & 22.99±0.57 & 34.59±0.45 & 1263.92±47.18 & 16.45±0.77 \\
\cmidrule(lr){2-7}
 & \multirow{2}{*}{PFL} & FedPer \cite{fedper} & 22.99±0.06 & 34.68±0.29 & 1272.55±6.34 & 16.59±0.20 \\
 &                    & pFedMe \cite{pfedme} & 21.40±0.30 & 32.77±0.52 & 1150.65±20.94 & 14.91±0.38 \\
\cmidrule(lr){2-7}
 & \multirow{2}{*}{TrafficFL} & FedTPS \cite{zhou2024traffic,zhou2025fedtps} & 21.83±0.42 & 33.49±0.79 & 1204.27±55.71 & 15.37±0.41 \\
 &                           & FedGCN \cite{hu2024fedgcn} & 22.92±0.39 & 34.86±0.26 & 1298.40±25.27 & 16.38±0.64 \\
\cmidrule(lr){2-7}
 & \multirow{3}{*}{Ours} & AutoFed   & \textbf{20.79±0.13} & \textbf{31.81±0.32} & \textbf{1084.61±5.91} & \textbf{14.72±0.49} \\
 &                      & w/o AE    & 20.76±0.07 & 31.90±0.31 & 1091.59±5.54 & 14.32±0.22 \\
 &                      & w/o FedBN & 21.94±0.26 & 33.47±0.47 & 1193.12±10.30 & 15.12±0.13 \\
\midrule \midrule
\multirow{9}{*}{PEMS08} & \multicolumn{2}{c}{Local Training} & 18.65±0.36 & 27.64±0.53 & 790.42±23.90 & 12.44±0.45 \\
\cmidrule(lr){2-7}
 & \multirow{2}{*}{FL} & FedAvg \cite{fedavg} & 19.29±0.42 & 28.78±0.38 & 860.93±26.91 & 13.04±0.51 \\
 &                    & FedProx \cite{fedprox} & 19.26±0.16 & 28.72±0.14 & 857.98±12.54 & 13.02±0.28 \\
\cmidrule(lr){2-7}
 & \multirow{2}{*}{PFL} & FedPer \cite{fedper} & 19.18±0.31 & 28.62±0.33 & 851.10±20.90 & 12.71±0.36 \\
 &                    & pFedMe \cite{pfedme} & 17.69±0.46 & 26.64±0.67 & 740.20±30.35 & 11.40±0.47 \\
\cmidrule(lr){2-7}
 & \multirow{2}{*}{TrafficFL} & FedTPS \cite{zhou2024traffic,zhou2025fedtps} & 17.97±0.41 & 27.05±0.64 & 764.11±29.99 & 11.81±0.21 \\
 &                           & FedGCN \cite{hu2024fedgcn} & 19.66±0.52 & 29.10±0.55 & 883.01±39.81 & 13.77±1.09 \\
\cmidrule(lr){2-7}
 & \multirow{3}{*}{Ours} & AutoFed   & \textbf{16.83±0.17} & \textbf{25.62±0.21} & \textbf{684.63±14.16} & \textbf{11.10±0.26} \\
 &                      & w/o AE    & 16.89±0.23 & 25.71±0.24 & 689.78±16.04 & 11.23±0.21 \\
 &                      & w/o FedBN & 17.96±0.33 & 27.16±0.44 & 769.56±26.49 & 11.80±0.39 \\
\bottomrule
\end{tabular}
\end{adjustbox}
\label{tab:expand}
\end{table*}

The experimental results for the TDP and TFP tasks are summarized in Tables~\ref{tab:exp} and~\ref{tab:expand}, respectively. We report the mean ± standard deviation over three independent runs.

In the TDP task, AutoFed achieves highly competitive performance across all four scenarios. In the most challenging scenario, S0 (whole-city Uber/Lyft as two clients), FedGCN obtains the best MAE/RMSE/MSE on Uber, while AutoFed delivers the best MAPE on Uber and dominates all four metrics on Lyft. In the more realistic regional scenarios (S1: five Uber clients; S2: five Lyft clients; S3: all ten clients), AutoFed consistently ranks first in almost every metric. According to the results, two important observations emerge: First, S0 yields the worst overall performance for all methods, confirming our design choice that a single city-wide graph introduces harmful interference for adaptive adjacency models such as AGCRN. Second, as client heterogeneity increases from S1/S2 to S3, many baselines suffer noticeable degradation, whereas AutoFed remains robust, consistent with the stability analysis in Appendix \ref{sec:theory} (Remark \ref{rem:hetero}), which shows that an outlier client perturbs the shared prompt generator only in proportion to its aggregation weight. The ablation studies further validate our design: removing either the AE denoiser or the FedBN-based adapter leads to consistent performance drops (especially visible in S0 and S3), demonstrating the contribution of both modules to effective denoise and non-IID adaptation.

In the TFP task, AutoFed establishes clear best results on the three PEMS benchmarks (averaged over client counts of 2, 4, 6, 8, and 10). Note that the results differ slightly from those reported in FedTPS \cite{zhou2024traffic, zhou2025fedtps} because, for fairness, we have removed practical tricks such as curriculum learning from their official repository (since this mechanism could be used for all methods or not), which are not part of the standard method itself and are not mentioned in the paper. The ablation variants again confirm that both the AE denoiser and the FedBN adapter are critical, where their removal leads to visible degradation. More detailed discussion and analysis, including training cost and complexity (Appendix \ref{sec:discussion}) and representation visualization (Appendix \ref{sec:visual}), are provided in the appendices.

\section{Conclusion}

In this paper, we propose AutoFed, a novel personalized federated traffic prediction framework that pairs a locally trained personalized predictor with a federated representor, which turns non-IID client data into a common global representation used as a prompt for the predictor's decoding. Because clients share this prompt rather than aggregated parameters, AutoFed supports road networks with differing numbers of nodes, keeps per-round communication independent of the graph size, and converges under standard assumptions. It also removes the dataset-specific manual design (hand-crafted graph construction and pattern-filter selection) that prior work must repeat per dataset, retaining only standard hyper-parameters that need no per-dataset redesign. Experiments on real-world travel demand and traffic flow prediction datasets show strong performance with reduced communication costs. More discussions are provided in Appendix \ref{sec:limitation}.

\bibliography{aaai2027}


\clearpage
\appendix



\section{Algorithm} \label{sec:alg}

The detailed process of our AutoFed is illustrated in Algorithm \ref{alg}.

\begin{algorithm}[htbp]
\caption{Training Process of AutoFed}
\label{alg}
\begin{algorithmic}[1]
\REQUIRE Number of clients $\mathcal{N}$, local datasets $\{D_i\}_{i=1}^\mathcal{N}$, shared model parameters $\Theta^0$, personal model parameters $\{\theta_i^0\}_{i=1}^\mathcal{N}$, communication rounds $M$, local epochs $E$
\ENSURE Personalized models $\{\theta_i^M, \Theta^M\}_{i=1}^\mathcal{N}$
\STATE \textbf{Server executes:}
\STATE Initialize shared parameters $\Theta^0$
\STATE Broadcast $\Theta^0$ to all clients
\FOR{each communication round $m = 0, 1, \dots, M-1$}
    \STATE Randomly select a subset of clients $S_m \subseteq \{1, \dots, \mathcal{N}\}$
    \FOR{each client $i \in S_m$ \textbf{in parallel}}
        \STATE $\theta_i^{m+1}, \Delta\Theta_i^{m} \leftarrow \text{ClientUpdate}(i, \theta_i^m, \Theta^m)$
    \ENDFOR
    \STATE Aggregate shared parameters: $\Theta^{m+1} \leftarrow \Theta^m + \frac{1}{|S_m|}\sum_{i \in S_m} \Delta\Theta_i^{m}$
    \STATE Broadcast $\Theta^{m+1}$ to all clients
\ENDFOR

\STATE
\STATE \textbf{ClientUpdate}(client $i$, personal parameters $\theta_i$, shared parameters $\Theta$):
\STATE Initialize local shared parameters: $\Theta_i \leftarrow \Theta$
\FOR{each local epoch $e = 1$ to $E$}
    \FOR{each batch $(X_b, Y_b) \in D_i$}
        \STATE \textbf{Forward Pass:}
        \STATE Compute denoised features: $p = \text{AE-enc}(X_b)$, $\hat{X}_b = \text{AE-dec}(p)$
        \STATE Extract local features: $p_l = \text{Encoder}(p)$
        \STATE Generate global prompt: $p_g = \text{Adapter}(p_l)$
        \STATE Generate predictions: $\hat{Y}_b = \text{PP}(X_b, p_g)$
        
        \STATE \textbf{Loss Computation:}
        \STATE Compute AE loss: $L^{ae} = \text{MAE}(X_b, \hat{X}_b)$
        \STATE Compute prediction loss: $L^{pre} = \text{MAE}(Y_b, \hat{Y}_b)$
        \STATE Compute adaptive weight: $\alpha = \frac{L^{ae}}{L^{pre}}$
        \STATE Compute total loss: $L = L^{pre} + \alpha \cdot L^{ae}$
        
        \STATE \textbf{Backward Pass:}
        \STATE Update $\theta_i$ and $\Theta_i$ via gradient descent: $\nabla_{\theta_i, \Theta_i} L$
    \ENDFOR
\ENDFOR
\STATE Compute parameter update: $\Delta\Theta_i = \Theta_i - \Theta$
\STATE Return $\theta_i$, $\Delta\Theta_i$
\end{algorithmic}
\end{algorithm}

\section{Introduction of Comparative Methods} \label{sec: Detailed Description of Comparative Methods}
\begin{itemize}[left=0pt]
    \item \textbf{FedAvg \cite{fedavg}:} This classical FL method enables each client to train on their individual data at each round, after which the server parameters are updated to the average of all clients' parameters, which are then distributed before the next round.
    
    \vspace{0.1cm}

    \item \textbf{FedProx \cite{fedprox}:} FedProx introduces a proximal term to the loss function of FedAvg, preventing the local model from deviating too far from the global model. However, it still aims to optimize a global model rather than personalized ones.
    
    \vspace{0.1cm}
    
    \item \textbf{FedPer \cite{fedper}:} As a classical PFL method, FedPer allows different clients to share a common base layer for robust feature extraction while maintaining personalized output layers for tailored results.
    
    \vspace{0.1cm}
    
    \item \textbf{pFedMe \cite{pfedme}:} pFedMe is a PFL framework that utilizes Moreau envelopes as a regularized loss function, enabling each client to efficiently find its optimal model while leveraging global information.
    
    \vspace{0.1cm}
    
    \item \textbf{FedTPS \cite{zhou2024traffic,zhou2025fedtps}:} Similar to our method, FedTPS employs a low-pass filter for stable pattern extraction and a pattern repository for prompt generation. However, these processes require manual settings and hyper-parameter tuning, and the aggregation in the pattern repository is based on similarity among clients, leading to higher computational costs.
    
    \vspace{0.1cm}
    
    \item \textbf{FedGCN \cite{hu2024fedgcn}:} FedGCN is based on the assumption that potential neighbor nodes of all real nodes exist within the graph. It introduces the MendGCN module to consider these external factors.
\end{itemize}

\section{Training Cost and Complexity Analysis} \label{sec:discussion}

We analyze the training efficiency of AutoFed from two complementary angles: an empirical comparison of the per-round cost against representative baselines, and an asymptotic analysis that explains the observed trends.

\begin{table}[htbp]
\centering
\caption{Training Costs Comparison}
\begin{tabular}{lcc}
\toprule
\multirow{2}{*}{Method} & Computation & Communication \\
& (time/round) (s) & (parameter/round) \\
\midrule
FedAvg \cite{fedavg} & \textbf{76.2} & 5.5M \\
FedProx \cite{fedprox} & 77.2 & 5.5M \\
\midrule
FedPer \cite{fedper} & 81.6 & 2.9M \\
pFedMe \cite{pfedme} & 90.0 & 5.5M \\
\midrule
 FedTPS \cite{zhou2024traffic,zhou2025fedtps} & 106.4 & \textbf{0.1K} \\
FedGCN \cite{hu2024fedgcn} & \underline{76.8} & 12.9M \\
\midrule
 AutoFed & 104.6 & \underline{175.0K} \\
\bottomrule
\end{tabular}
\label{tab:cost}
\end{table}

\begin{figure*}[htbp]
\centering
\includegraphics[width=\textwidth]{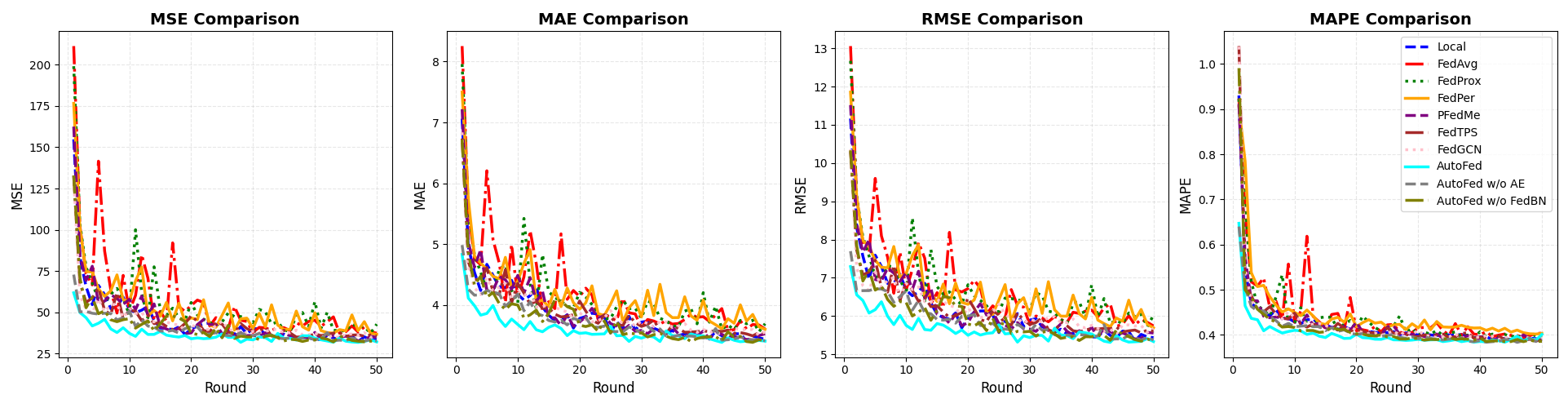}
\caption{Training Process: This figure shows the change of different metrics in valid set during training.}
\label{fig:train}
\end{figure*}

\subsection{Empirical Training Costs}

Using scenario S1 of the TDP task as a running example, we compare the convergence speed, computational cost, and communication cost during training. Fig.~\ref{fig:train} shows how the validation metrics evolve as training proceeds, where AutoFed converges faster than the baselines. Table~\ref{tab:cost} reports the detailed per-round costs. AutoFed and FedTPS have the lowest communication costs but the highest computation times, since both generate a prompt at each round. Notably, AutoFed reaches a communication cost close to that of FedTPS while achieving markedly better accuracy across tasks (Tables~\ref{tab:exp} and~\ref{tab:expand}).

\subsection{Computation and Communication Complexity}

We now complement Table~\ref{tab:cost} with an asymptotic analysis that explains its two trends: AutoFed has a computation cost of the same order as a single AGCRN backbone, while its communication cost is independent of the graph size. We use the notation of Section~\ref{sec:methodology}: $n$ is the number of nodes of a client, $K$ the input feature dimension, $k$ the hidden and prompt dimension, $\mathcal T$ and $\mathscr T$ the history and horizon lengths, $L$ the number of AGCRN layers, and $E$ the number of local epochs per round.

\paragraph{Computation.}
The dominant cost of the backbone is the adaptive graph convolution inside AGCRN. Forming the adaptive adjacency $\tilde A$ from the learnable node embedding costs $O(n^2 k)$, and each graph convolution over the dense $n \times n$ adjacency costs $O(n^2 k + n k^2)$, where the $O(n^2 k)$ term dominates because $n \gg k$ in traffic graphs. A GRU cell applies three such convolutions, so one AGCRN pass over a length-$T$ sequence with $L$ layers costs $O(T L n^2 k)$. AutoFed performs three such passes per sample, namely the PP encoder at $O(\mathcal T L n^2 k)$, the PP decoder at $O(\mathscr T L n^2 k)$, and the FR graph encoder at $O(\mathcal T L n^2 k)$. The remaining FR modules act on already compressed features and are cheaper: the AE denoiser applies a per-node feature map at a cost of $O(\mathcal T n K k)$, and the client-aligned adapter applies a shared per-node linear map to the single matrix $p_l \in \mathbb R^{n \times k}$ produced by the FR encoder, at a cost of $O(n k^2)$. The per-sample forward cost is therefore
\begin{equation}
O\big((2\mathcal T + \mathscr T)\,L\,n^2 k \;+\; \mathcal T n K k \;+\; n k^2\big),
\label{eq:comp}
\end{equation}
and one training round over client $i$ with $E$ local epochs multiplies this by $E N_i$. The adapter term $O(n k^2)$ carries no sequence length and is linear rather than quadratic in $n$, so it is dominated by the graph-convolution term whenever $k \ll \mathcal T L n$, which holds throughout our setting. A plain AGCRN predictor, as used by FedAvg-style baselines, already costs $O((\mathcal T + \mathscr T)L n^2 k)$, so the modules AutoFed adds, namely the FR encoder, the denoiser, and the adapter, raise the cost by less than a factor of two and leave the asymptotic order unchanged. The $n^2$ dependence stems from the dense adaptive adjacency and is inherent to AGCRN rather than introduced by AutoFed. This is consistent with Table~\ref{tab:cost}, where AutoFed's per-round time of $104.6$s is a small constant factor above FedAvg at $76.2$s.

\paragraph{Communication.}
Only the shared block $\Theta$ is exchanged each round. The following observation makes precise why this cost does not grow with the graph.

\begin{proposition}[Node-count-independent communication]\label{prop:comm}
Per round, each client transmits $O(|\Theta|)$ parameters, where the shared block satisfies $|\Theta| = O(k^2 + Kk)$. In particular it is independent of the node count $n$, the sequence lengths $\mathcal T$ and $\mathscr T$, and the size of the personalized predictor.
\end{proposition}

\begin{proof}
The shared block $\Theta$ consists of the AE denoiser and the shared linear layers of the adapter (Section~\ref{sec:methodology}); the AGCRN predictor, the personalized FR encoder, and all batch normalization layers are private and are never communicated. Because $\Theta$ is shared across clients whose graphs have different node counts, its layers cannot act on the node axis and must act on the per-node feature dimension instead: the denoiser maps the feature dimension $K$ to $k$ and back using $O(Kk)$ parameters, and the shared adapter linear maps $k$ to $k$ using $O(k^2)$ parameters. Neither depends on $n$ or on the sequence lengths, which enter only as the sizes of the axes over which the layers are broadcast. Summing gives $|\Theta| = O(k^2 + Kk)$. Each round uploads the update $\Delta\Theta_i$ and downloads $\Theta$, both of this size.
\end{proof}

By contrast, aggregating a full model as in FedAvg or FedGCN transmits the predictor as well, whose AGCRN node embedding alone is $O(nk)$ and therefore grows with the graph. This is the source of the large gap in Table~\ref{tab:cost}: AutoFed communicates $175.0$K parameters per round, against $5.5$M for FedAvg and $12.9$M for FedGCN, a reduction of roughly $30$ to $70$ times, and the gap widens as the number of nodes grows. AutoFed thus trades a modest constant-factor increase in local computation for a communication cost that is asymptotically smaller and, importantly, independent of the graph size. This is the regime that matters in cross-silo traffic prediction, where the bandwidth between agencies is the binding constraint while each silo has ample local compute.

\section{Representation Visualization} \label{sec:visual}

In this section we analyze the prompt produced by the federated representor in order to understand what the client-aligned adapter learns and why the prompt helps prediction. We organize the analysis around three questions: how the adapter reshapes the prompt space across clients, whether the prompt is causally relied upon by the decoder, and what traffic information it encodes. Throughout, we distinguish the local prompt $p_l$, the output of the low-frequency encoder before the client-aligned adapter, from the global prompt $p_g$, the output of the adapter that is fed to the decoder as the prefix token. Unless stated otherwise, we capture $p_l$ and $p_g$ with forward hooks on held-out test windows so the trained models are left untouched, and every row of $p_l$ or $p_g$ corresponds to one node in one input window. We report results on both the TDP task (scenarios S0 and S1) and the TFP task (PEMS04 and PEMS08 with four clients), and use scenario S1 of the TDP task as a running example when a single setting is shown.

\subsection{How the Adapter Reshapes the Prompt Space}
\label{sec:visual_align}

\begin{figure}[t!]
\centering
\subfloat[\footnotesize{Client Distribution under Local Representation} \label{fig:local_proportion}]{\includegraphics[width=0.5\textwidth]{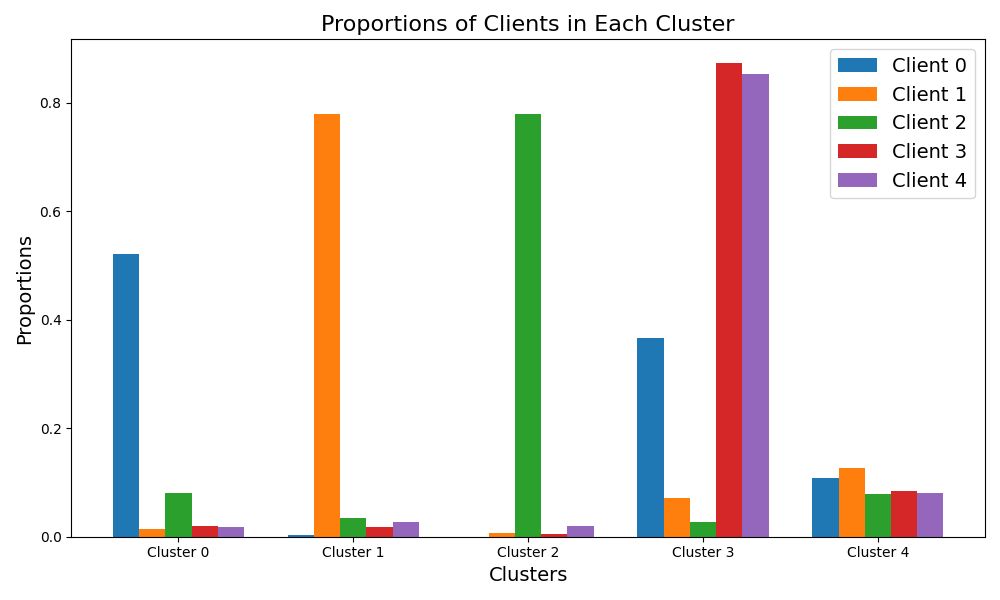}} \\
\subfloat[\footnotesize{Client Distribution under Global Representation} \label{fig:global_proportion}]{\includegraphics[width=0.5\textwidth]{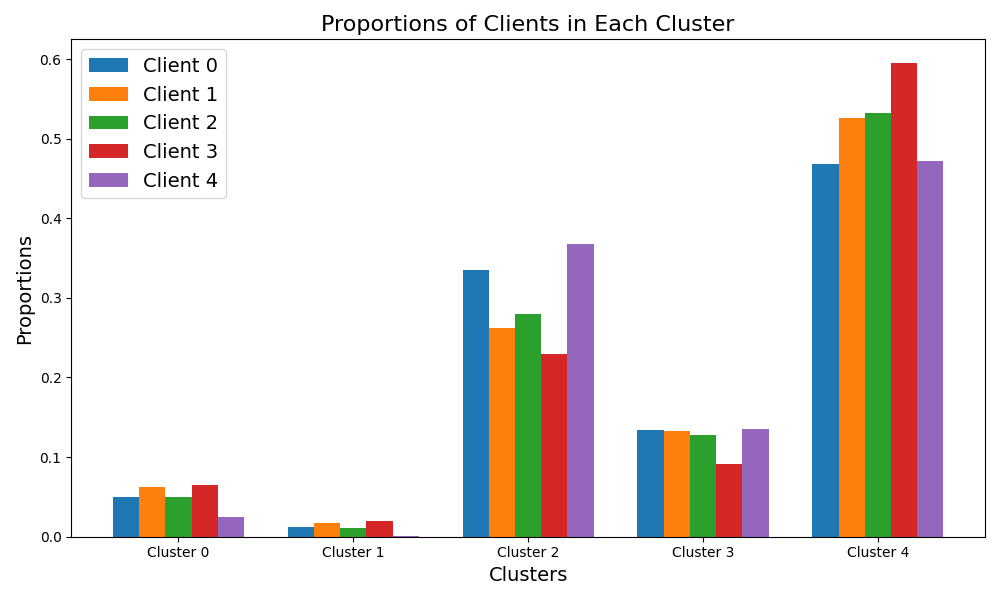}}
\caption{Comparison Between Local Representation and Global Representation}
\label{fig:visual}
\end{figure}

\begin{figure}[t!]
\centering
\subfloat[\footnotesize{TFP, PEMS04 with four clients} \label{fig:align_tfp}]{\includegraphics[width=0.48\textwidth]{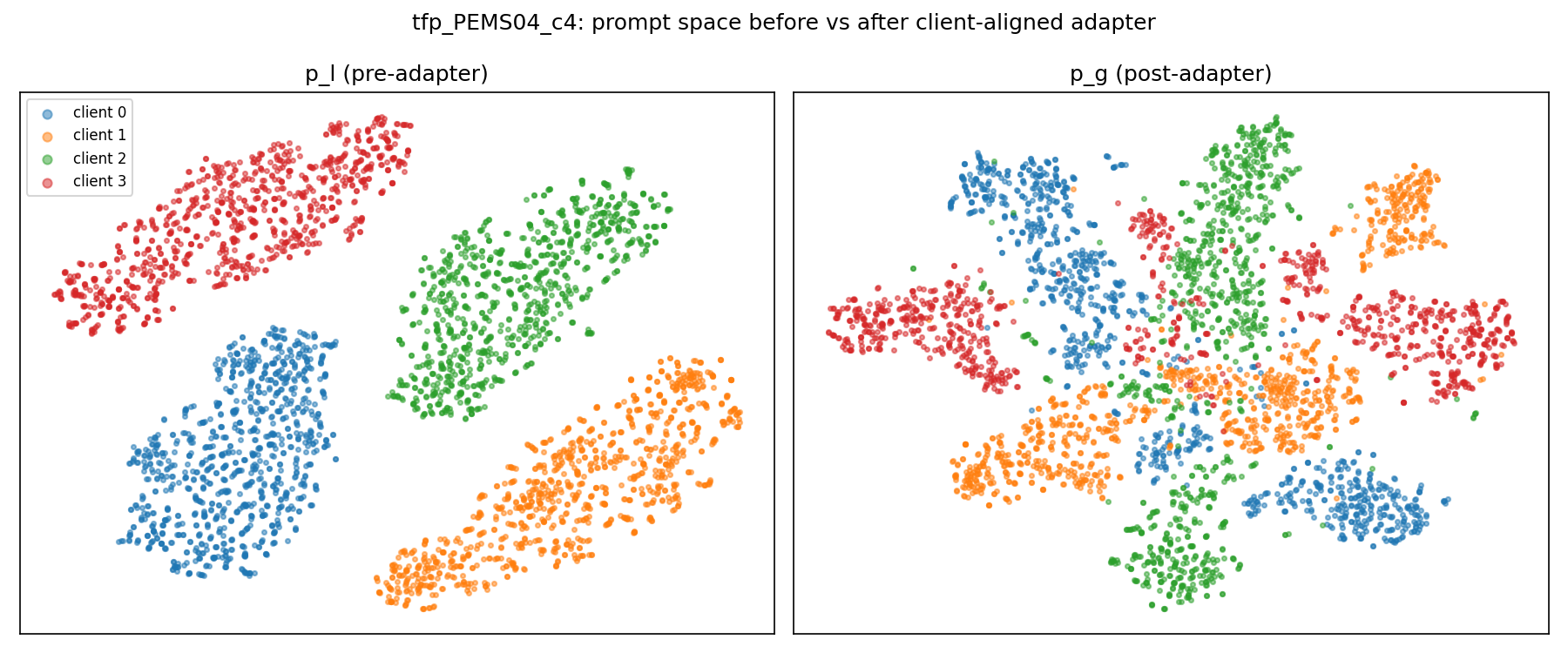}} \\
\subfloat[\footnotesize{TDP, scenario S0} \label{fig:align_tdp}]{\includegraphics[width=0.48\textwidth]{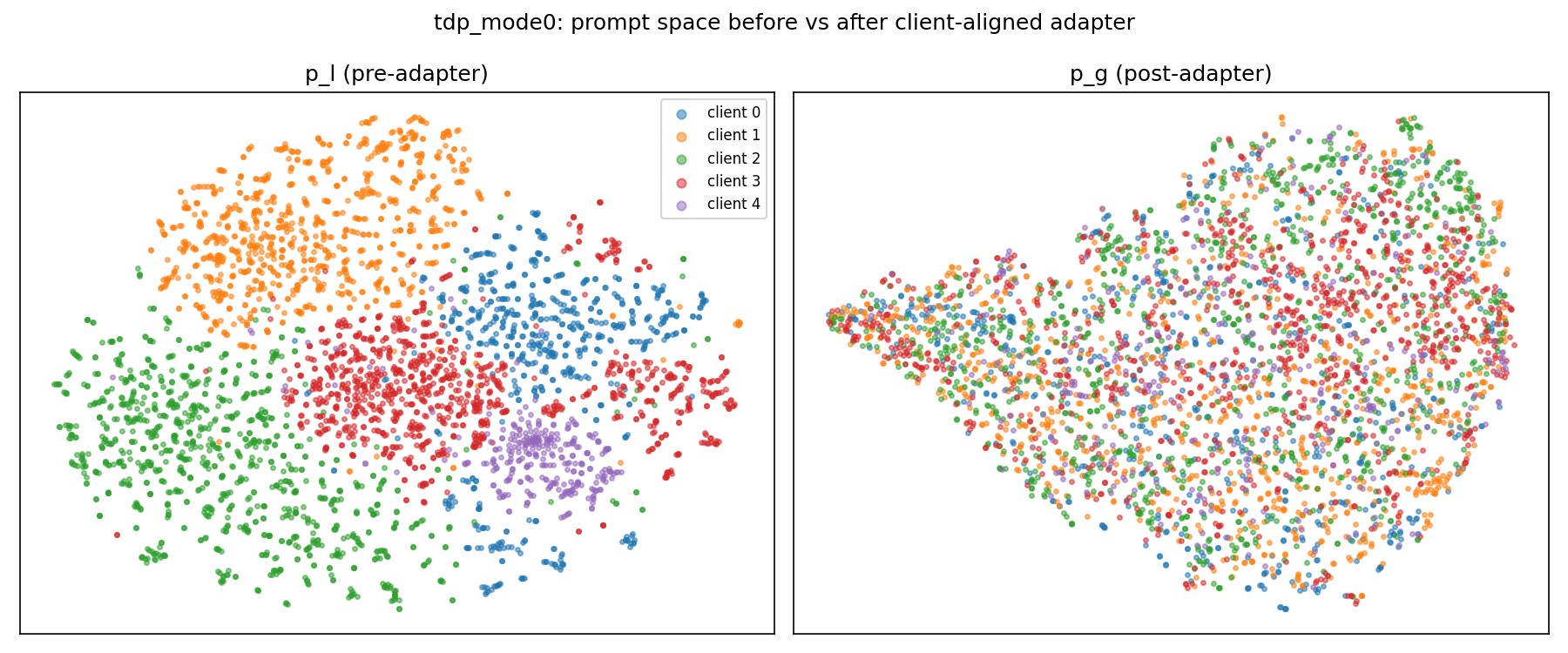}}
\caption{Two-dimensional t-SNE embeddings of the prompt before ($p_l$, left) and after ($p_g$, right) the client-aligned adapter, colored by client. Local prompts form clean per-client clusters, whereas the adapter relocates clients into a shared, overlapping space.}
\label{fig:align}
\end{figure}

We first ask how the client-aligned adapter reorganizes the prompt across clients, using two complementary views of the same phenomenon. The first is a clustering view (Fig.~\ref{fig:visual}): we combine all $p_l$ and $p_g$ across clients, apply KMeans clustering initialized at the per-client centers, and measure the proportion of each client within every cluster. Under the local prompt, clients differ substantially and their features concentrate in different clusters, whereas under the global prompt each cluster becomes a relatively uniform mixture of clients, indicating that the adapter suppresses client-specific characteristics and surfaces globally shared traffic patterns. The second is a projection view (Fig.~\ref{fig:align}): we embed $p_l$ and $p_g$ into two dimensions with t-SNE (PCA to $50$ dimensions beforehand) and color each point by its client of origin. The two views agree. Before the adapter, the local prompts of different clients occupy clearly separated regions, reflecting the heterogeneous statistics of each region; after the adapter, the clusters are pulled together and interleave heavily, so a point's client can no longer be read off from its position.

We quantify this alignment with the silhouette score using the client label as the cluster assignment, where a higher value means cleaner per-client separation. The score drops consistently from $p_l$ to $p_g$ on every setting, from $0.162$ to $0.003$ on PEMS04 and from $0.098$ to $0.013$ on PEMS08. On the two TDP scenarios the score is already negative for $p_l$ (about $-0.02$ to $-0.05$) and becomes more negative for $p_g$; here the clients overlap in prompt space even before the adapter because of how the demand data is partitioned, so for TDP the silhouette is a weak indicator and we rely instead on the geometric measure below. To check that the relocation preserves the internal geometry of the representation rather than destroying it, we compute the linear centered kernel alignment (CKA) between $p_l$ and $p_g$ per client. On the TFP task the CKA is high ($0.92$ to $0.96$), showing that the adapter mainly repositions each client as a whole while keeping the relative structure of its samples intact; on the TDP task it is lower and more variable ($0.31$ to $0.76$), indicating a stronger rewrite for the more heterogeneous demand data. In both cases the adapter behaves as its name suggests: it projects heterogeneous local representations into a common coordinate system in which cross-client knowledge can be shared, rather than sharpening client identity.

\subsection{Is the Prompt Causally Used?}
\label{sec:visual_interv}

\begin{figure}[t!]
\centering
\includegraphics[width=0.48\textwidth]{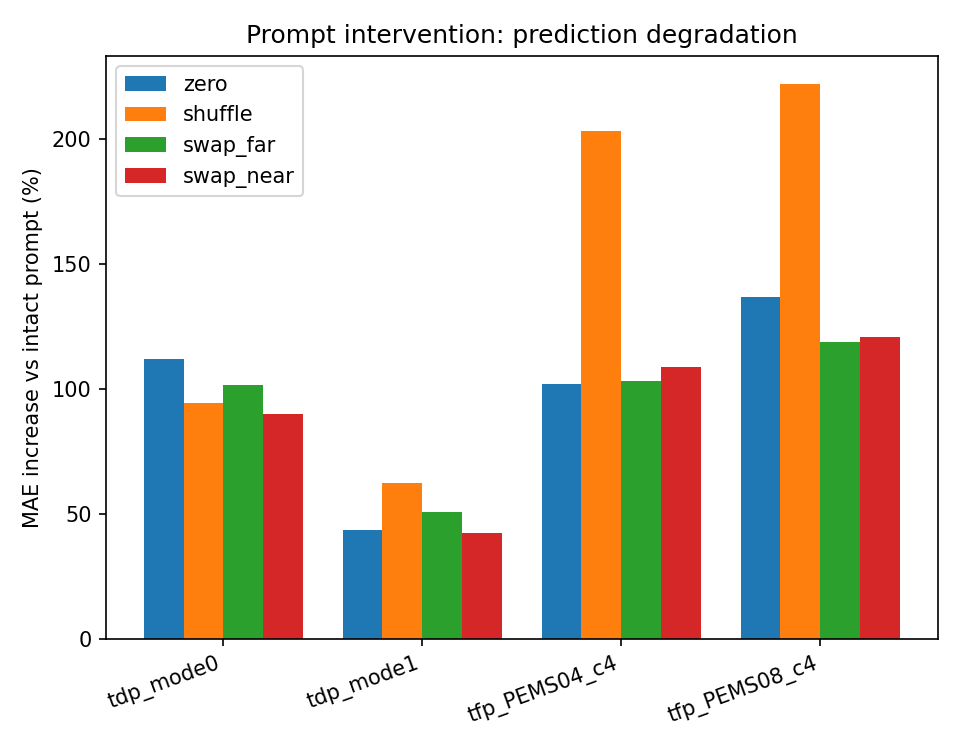}
\caption{Relative increase in test MAE when the prompt $p_g$ is corrupted at inference, averaged over clients. Every intervention degrades accuracy, and shuffling the prompt across samples is the most damaging, which shows the prompt carries per-window and per-node information rather than a coarse client label.}
\label{fig:intervention}
\end{figure}

\begin{table}[t!]
\centering
\small
\setlength{\tabcolsep}{4pt}
\begin{tabular}{lcccc}
\toprule
Setting & \texttt{zero} & \texttt{shuffle} & \texttt{swap-far} & \texttt{swap-near} \\
\midrule
TDP S0        & $+112$ & $+95$  & $+102$ & $+90$  \\
TDP S1        & $+43$  & $+62$  & $+51$  & $+42$  \\
TFP PEMS04    & $+102$ & $+203$ & $+103$ & $+109$ \\
TFP PEMS08    & $+137$ & $+222$ & $+119$ & $+121$ \\
\bottomrule
\end{tabular}
\caption{Relative test MAE increase (\%) under prompt intervention, averaged over clients. Errors are measured on each task's native scale through the per-client scaler (original flow units for TFP and normalized demand for TDP), so magnitudes are comparable within a task but not necessarily across tasks.}
\label{tab:intervention}
\end{table}

A visualization on its own cannot tell us whether the prompt is actually relied upon by the model. We therefore run a causal intervention: at inference we overwrite the adapter output $p_g$ and measure how much the test MAE rises relative to the intact model (Fig.~\ref{fig:intervention}, Table~\ref{tab:intervention}). We consider four corruptions. \texttt{zero} replaces the prompt with zeros, ablating it entirely; \texttt{shuffle} permutes the prompt rows within a batch, keeping the value distribution but breaking the correspondence between a prompt and its own window and node; \texttt{swap-far} and \texttt{swap-near} replace every row with the mean prompt of the most and least distant donor client, where distance is the Euclidean distance between per-client mean prompts. Across all settings every corruption inflates the error substantially, by $43\%$ to more than $200\%$, confirming that the prompt is genuinely consumed by the decoder rather than acting as a passive side input. The pattern across corruptions is informative. \texttt{shuffle} is consistently the most harmful, especially on the TFP task where it roughly triples the error, which means the prompt is not a single per-client signal but carries information tied to each specific window and node. In contrast, replacing the prompt with a static donor vector (\texttt{swap}) hurts about as much as zeroing it and no more, indicating that a frozen city embedding is nearly worthless and that the value of the prompt lies in its input-conditioned variation. This directly supports the design choice of generating the prompt dynamically from each input rather than storing one vector per client.

Notably, the distance between clients in prompt space does not predict transferability: \texttt{swap-far} and \texttt{swap-near} cause almost the same degradation, and the nearer donor is sometimes worse. Rather than a limitation, this reinforces the point above. What matters is not which city a prompt is borrowed from but whether it is dynamically matched to the current input, so a static prompt is damaging regardless of its source.

\subsection{What Does the Prompt Encode?}
\label{sec:visual_probe}

\begin{table}[t!]
\centering
\small
\setlength{\tabcolsep}{5pt}
\begin{tabular}{lcccc}
\toprule
& \multicolumn{2}{c}{Level ($R^2$)} & \multicolumn{2}{c}{Volatility ($R^2$)} \\
\cmidrule(lr){2-3} \cmidrule(lr){4-5}
Setting & probe & control & probe & control \\
\midrule
TDP S0     & $0.10$ & $0.00$ & $0.13$ & $0.00$ \\
TDP S1     & $0.14$ & $0.00$ & $0.20$ & $0.00$ \\
TFP PEMS04 & $0.87$ & $0.01$ & $0.51$ & $0.00$ \\
TFP PEMS08 & $0.79$ & $0.00$ & $0.35$ & $0.00$ \\
\bottomrule
\end{tabular}
\caption{Linear probing of $p_g$. We report the test $R^2$ of a ridge regressor predicting the within-window mean (level) and standard deviation (volatility) of the primary channel, against a shuffled-label control.}
\label{tab:probe}
\end{table}

Finally we ask what traffic content the prompt carries. We fit a ridge regressor from $p_g$ to two simple window statistics of the primary channel, its mean, which reflects the current traffic level, and its standard deviation, which reflects short-term volatility, and report the test $R^2$ against a control trained on shuffled labels (Table~\ref{tab:probe}). If the prompt encodes these quantities linearly, the probe should score well above the control, whose $R^2$ is near zero by construction. On the TFP task the current traffic level is decoded strongly ($R^2$ of $0.87$ and $0.79$) and volatility moderately ($0.51$ and $0.35$), so the flow prompt clearly represents how busy and how bursty the current window is. On the TDP task the same probe is weak ($R^2$ of $0.10$ to $0.20$) but still far above the control, which is expected because the demand input has nine channels and the mean of a single channel is not its dominant signal. Overall the prompt is not an opaque code: it linearly exposes interpretable traffic quantities, most strongly on the flow task where such statistics are the primary signal.

Taken together, the three analyses give a concrete answer to what the prompt is and why it helps. The adapter aligns heterogeneous local prompts into a shared space (Figs.~\ref{fig:visual} and~\ref{fig:align}), the resulting prompt is causally relied upon and must be matched to each specific input rather than fixed per client (Table~\ref{tab:intervention}), and it encodes interpretable traffic level and volatility (Table~\ref{tab:probe}). The evidence is strongest and most consistent on the TFP task, where flow is a low-dimensional, strongly periodic signal: alignment is clean, the prompt is heavily used, and its content is decoded with high fidelity. On the TDP task the same trends hold in direction but are weaker, because travel demand is a higher-dimensional and more diffuse signal whose informative structure is not captured by a single-channel statistic; there the story rests mainly on the alignment and causal-usage evidence rather than on the linear probe. In both regimes, however, the prompt turns out to be an input-conditioned, causally used, and partially interpretable component rather than an opaque code.

\section{Theory Analysis} \label{sec:theory}

This appendix collects the theoretical analysis of AutoFed. The first part, \emph{Convergence and Stability of Federated Training}, is fully rigorous. It analyzes the training procedure of Algorithm~\ref{alg} as an instance of partial model personalization and shows that, under standard smoothness and bounded variance assumptions, the iterates converge to a stationary point of the personalized objective in Eq.~\eqref{eq:target} at the usual nonconvex rate and cannot diverge for a suitable step size. We also prove that the objective value attainable by AutoFed is never worse than that of a fully shared, FedAvg style model, which formally answers whether prompt based personalization can degrade to plain federated averaging, and we explain why a single outlier client cannot destabilize training.

The second part, \emph{Interpreting the Prompt}, is deliberately interpretive rather than fully rigorous. It gives a probabilistic reading of the prompt as a global conditioning signal, a domain adaptation heuristic for why aligning clients in the prompt space aids generalization, and a low rank view of prompt conditioning. These arguments build intuition and are consistent with our experiments, but they rely on informal steps that we flag explicitly. Their rigorous versions, in particular a quantitative contraction factor for the cross client divergence, are left for future work.

Throughout, $z = (\Theta, \theta_1, \dots, \theta_{\mathcal N})$ denotes the full parameter vector, where $\Theta$ collects the shared parameters (the AE denoiser and the shared linear layers of the adapter) and $\theta_i$ collects the parameters kept private to client $i$ (its AGCRN predictor, personalized encoder, and batch normalization layers). We write the objective of Eq.~\eqref{eq:target} as
\begin{equation}
F(z) = \sum_{i=1}^{\mathcal N} w_i F_i(\Theta, \theta_i), \qquad w_i = \frac{|V_i|}{\sum_{j=1}^{\mathcal N}|V_j|},
\label{eq:obj-theory}
\end{equation}
where $F_i(\Theta,\theta_i) = \mathbb{E}_{\xi\sim\mathcal D_i}[\ell_i(\Theta,\theta_i;\xi)]$ is the expected local loss of client $i$ and $\ell_i \ge 0$ is the per sample MAE loss of Eq.~\eqref{eq:mae}.

\subsection{Convergence and Stability of Federated Training}

Because each private block $\theta_i$ appears in only the $i$-th summand of Eq.~\eqref{eq:obj-theory}, the block gradients are
\begin{equation}
\nabla_\Theta F = \sum_{i=1}^{\mathcal N} w_i \nabla_\Theta F_i,
\qquad
\nabla_{\theta_i} F = w_i \nabla_{\theta_i} F_i .
\label{eq:blockgrad}
\end{equation}
This is exactly the structure of \emph{partial model personalization}~\cite{pillutla2022federated}: a shared block $\Theta$ that is aggregated across clients and private blocks $\theta_i$ that are updated only locally. AutoFed's training in Algorithm~\ref{alg} is the simultaneous variant, in which both blocks are updated in the same local pass and only $\Theta$ is communicated. We make the following standard assumptions.

\begin{assumption}[Smoothness]\label{ass:smooth}
Each $F_i$ is continuously differentiable and $L$-smooth in $z$, that is $\|\nabla F_i(z) - \nabla F_i(z')\| \le L\|z - z'\|$ for all $z, z'$.
\end{assumption}

\begin{assumption}[Lower bound]\label{ass:lb}
$F$ is bounded below by some $F^\star > -\infty$. This holds here because $\ell_i \ge 0$ implies $F \ge 0$.
\end{assumption}

\begin{assumption}[Unbiased gradients with bounded variance]\label{ass:var}
Each client's stochastic block gradient $g_{i,\bullet}$, computed on a sampled mini batch, is unbiased with $\mathbb{E}[g_{i,\bullet}\mid z] = \nabla_\bullet F_i(z)$ and $\mathbb{E}\|g_{i,\bullet} - \nabla_\bullet F_i\|^2 \le \sigma^2$, and the noises of different clients are mutually independent.
\end{assumption}

We first analyze the transparent regime of full participation with one local step per round; the general case is discussed afterward. In this regime one round performs, with step size $\eta > 0$,
\begin{equation}
\Theta^{m+1} = \Theta^m - \eta \sum_{i} w_i\, g^m_{i,\Theta},
\qquad
\theta_i^{m+1} = \theta_i^m - \eta\, g^m_{i,\theta},
\label{eq:update}
\end{equation}
where the shared update aggregates client contributions with the objective weights $w_i$. Uniform averaging as written in Algorithm~\ref{alg} is the special case $w_i = 1/\mathcal N$, for which every result below holds verbatim with the uniformly weighted objective. Let $h^m$ denote the stacked update direction in Eq.~\eqref{eq:update}. Using Eq.~\eqref{eq:blockgrad}, a short computation gives $\mathbb{E}[h^m\mid z^m] = P\,\nabla F(z^m)$, where $P = \mathrm{blockdiag}(I,\, w_1^{-1} I, \dots, w_{\mathcal N}^{-1} I)$ is a fixed positive definite preconditioner with $\lambda_{\min}(P) = 1$ and $\lambda_{\max}(P) = 1/w_{\min}$, writing $w_{\min} := \min_i w_i$.

\begin{proposition}[Descent and convergence]\label{prop:converge}
Under Assumptions~\ref{ass:smooth} to \ref{ass:var}, if the step size satisfies $\eta \le w_{\min}/L$, then the iterates of Eq.~\eqref{eq:update} obey, for every round $m$,
\begin{equation}
\mathbb{E}[F(z^{m+1})\mid z^m] \le F(z^m) - \frac{\eta}{2}\|\nabla F(z^m)\|^2 + \frac{L\eta^2}{2}V,
\label{eq:descent}
\end{equation}
with $V := (\mathcal N + 1)\sigma^2$. Averaging over $M$ rounds,
\begin{equation}
\frac{1}{M}\sum_{m=0}^{M-1}\mathbb{E}\|\nabla F(z^m)\|^2 \le \frac{2\big(F(z^0) - F^\star\big)}{\eta M} + L\eta V.
\label{eq:rate}
\end{equation}
In particular, the choice $\eta = \min\{\,w_{\min}/L,\ \sqrt{2(F(z^0)-F^\star)/(LVM)}\,\}$ gives $\min_{m<M}\mathbb{E}\|\nabla F(z^m)\|^2 = O(1/\sqrt{M})$, and in the deterministic case $\sigma = 0$ the objective decreases monotonically to a stationary point at rate $O(1/M)$.
\end{proposition}

\begin{proof}
Fix a round $m$ and condition on $z^m$. Each $F_i$ is $L$-smooth in $z$ by Assumption~\ref{ass:smooth}, so their convex combination $F$ is also $L$-smooth, and the descent lemma gives
\[
F(z^{m+1}) \le F(z^m) + \langle \nabla F(z^m), z^{m+1}-z^m\rangle + \tfrac{L}{2}\|z^{m+1}-z^m\|^2 .
\]
Since $z^{m+1}-z^m = -\eta h^m$ and $\mathbb{E}[h^m\mid z^m] = P\nabla F(z^m)$, taking the conditional expectation yields
\[
\mathbb{E}[F(z^{m+1})\mid z^m] \le F(z^m) - \eta\langle \nabla F, P\nabla F\rangle + \tfrac{L\eta^2}{2}\,\mathbb{E}\|h^m\|^2 .
\]
Because $\mathbb{E}[h^m] = P\nabla F$, we have $\mathbb{E}\|h^m\|^2 = \|P\nabla F\|^2 + \mathbb{E}\|h^m - P\nabla F\|^2$. Under Assumption~\ref{ass:var} and the independence of client noise, the shared block variance is $\mathbb{E}\|\sum_i w_i(g_{i,\Theta}-\nabla_\Theta F_i)\|^2 = \sum_i w_i^2\,\mathbb{E}\|g_{i,\Theta}-\nabla_\Theta F_i\|^2 \le \sigma^2\sum_i w_i^2 \le \sigma^2$, using $\sum_i w_i^2 \le \max_i w_i \le 1$, while the $\mathcal N$ private blocks each contribute at most $\sigma^2$, so $\mathbb{E}\|h^m - P\nabla F\|^2 \le (\mathcal N+1)\sigma^2 = V$. For the two deterministic terms, expand $\langle \nabla F, P\nabla F\rangle = \|\nabla_\Theta F\|^2 + \sum_i w_i^{-1}\|\nabla_{\theta_i}F\|^2$ and $\|P\nabla F\|^2 = \|\nabla_\Theta F\|^2 + \sum_i w_i^{-2}\|\nabla_{\theta_i}F\|^2$. Since $w_i^{-1} \le w_{\min}^{-1}$ for every $i$, each coefficient obeys $w_i^{-2} \le w_{\min}^{-1} w_i^{-1}$, hence $\|P\nabla F\|^2 \le w_{\min}^{-1}\langle \nabla F, P\nabla F\rangle$. Substituting,
\[
\mathbb{E}[F(z^{m+1})\mid z^m] \le F(z^m) - \eta\Big(1 - \tfrac{L\eta}{2 w_{\min}}\Big)\langle\nabla F, P\nabla F\rangle + \tfrac{L\eta^2}{2}V .
\]
The condition $\eta \le w_{\min}/L$ makes $1 - \tfrac{L\eta}{2w_{\min}} \ge \tfrac12$, and $\langle \nabla F, P\nabla F\rangle \ge \lambda_{\min}(P)\|\nabla F\|^2 = \|\nabla F\|^2$, which gives Eq.~\eqref{eq:descent}. Rearranging Eq.~\eqref{eq:descent}, summing over $m = 0,\dots,M-1$, taking total expectation, and telescoping with Assumption~\ref{ass:lb} gives $\tfrac{\eta}{2}\sum_m \mathbb{E}\|\nabla F(z^m)\|^2 \le F(z^0)-F^\star + \tfrac{L\eta^2}{2}VM$; dividing by $\eta M/2$ yields Eq.~\eqref{eq:rate}. The stated step size balances the two terms of Eq.~\eqref{eq:rate} and gives the $O(1/\sqrt M)$ bound; when $\sigma=0$ we have $V=0$, so Eq.~\eqref{eq:descent} is a monotone decrease and $\sum_m\|\nabla F(z^m)\|^2 \le 2(F(z^0)-F^\star)/\eta$, that is $O(1/M)$.
\end{proof}

Proposition~\ref{prop:converge} answers the reviewers' stability concern directly: for any step size below $w_{\min}/L$, the expected objective decreases every round up to a noise floor of order $\eta^2$, so the training cannot diverge, and the averaged squared gradient vanishes at the standard nonconvex rate. The step size ceiling $w_{\min}/L$ also makes precise the intuitive fact that severely imbalanced clients, for which $w_{\min}$ is small, require a smaller learning rate for the same guarantee.

\begin{corollary}[No worse than federated averaging]\label{cor:fedavg}
Let $F^\star_{\mathrm{pers}} = \min_{\Theta,\{\theta_i\}} F(\Theta,\{\theta_i\})$ be the optimal value of the personalized objective in Eq.~\eqref{eq:obj-theory}, and let $F^\star_{\mathrm{shared}} = \min_{\Theta,\theta}\sum_i w_i F_i(\Theta,\theta)$ be the optimal value when all clients are forced to share a single predictor, as in FedAvg. Then $F^\star_{\mathrm{pers}} \le F^\star_{\mathrm{shared}}$.
\end{corollary}

\begin{proof}
The shared model is the restriction of Eq.~\eqref{eq:obj-theory} to the subspace $\{\theta_1 = \dots = \theta_{\mathcal N}\}$. Minimizing over a superset can only lower the minimum, hence $F^\star_{\mathrm{pers}}\le F^\star_{\mathrm{shared}}$.
\end{proof}

\begin{remark}[Degradation to FedAvg and outlier clients]\label{rem:hetero}
Corollary~\ref{cor:fedavg} shows that introducing the prompt based personalized components cannot raise the attainable training objective above that of plain federated averaging, so in the worst case AutoFed matches FedAvg rather than falling below it. This is the precise sense in which prompt aggregation does not degrade the framework. In addition, a single client $j$ whose distribution differs sharply from the rest enters the shared update in Eq.~\eqref{eq:update} only through the term $w_j\, g_{j,\Theta}$, whose norm is $O(w_j)$ under Assumption~\ref{ass:smooth} on a bounded parameter domain. The client's idiosyncrasies are otherwise absorbed by its fully private block $\theta_j$, namely its AGCRN predictor and batch normalization statistics, and never enter the private updates of other clients. An outlier therefore perturbs the shared prompt generator only in proportion to its aggregation weight and cannot by itself destabilize training. We note that this is a statement about optima and stationary convergence rather than global optimality, since $F$ is nonconvex.
\end{remark}

\paragraph{General case.}
The analysis above isolates the essential mechanism in the one local step, full participation setting. With $K > 1$ local steps and partial client participation, the same argument carries through up to an additional client drift term of order $O(\eta^2 K^2(\sigma^2 + \zeta^2))$, where $\zeta^2$ bounds the cross client diversity of the shared block gradient, $\sum_i w_i\|\nabla_\Theta F_i - \nabla_\Theta F\|^2 \le \zeta^2$ evaluated at the personalized parameters. This recovers the convergence guarantees established for partial model personalization~\cite{pillutla2022federated}, whose rate matches standard nonconvex FedAvg, so personalization does not slow the convergence of the shared block. A favorable and distinctive feature is that here $\zeta^2$ is measured at the personalized points $\theta_i$: because each client keeps a private predictor and private normalization, much of the heterogeneity that inflates $\zeta^2$ in fully shared FL is absorbed locally, and the effective diversity seen by the shared prompt generator is smaller than in FedAvg.

\subsection{Prompt as a Global Conditioning Signal}

We interpret the prompt matrix $p_g \in \mathbb{R}^{n \times k}$ generated by the Federated Representor (FR) as a compact, globally aligned representation that encodes shared spatiotemporal patterns across clients. For a given client $i$, the predictive distribution satisfies the following factorization by construction of the model:

\begin{equation}
P(\hat{y} \mid x) = P\bigl(\hat{y} \mid H_{\mathcal{T}}(x), p_g(x)\bigr),
\label{eq:predictive}
\end{equation}

where $H_{\mathcal{T}}(x) = \text{Encoder}(x; \mathbf{0})$ is the final encoder hidden state extracted from the raw client-specific input $x$, and $p_g(x) = \text{FR}(x)$ denotes the deterministic prompt generated by the Federated Representor. The decoder operates in an auto-regressive manner, conditioned on both $H_{\mathcal{T}}(x)$ and $p_g(x)$ (as the initial token).

This factorization explicitly separates two complementary roles: $H_{\mathcal{T}}(x)$ captures fine-grained, {client-local} spatiotemporal dynamics via the adaptive graph structure of AGCRN (its parameters are personalized per client), while $p_g(x)$ injects a {globally shared} task signal that summarizes recurring traffic patterns (e.g., daily/weekly rhythms) across heterogeneous regions. Thus, even though the raw data distributions $P_i(x)$ vary across clients, the shared prompt $p_g$ provides a common prior that the personalized decoder can condition on, reducing the burden of learning cross-client patterns from scratch.

\subsection{Generalization Advantage Under Client Heterogeneity}

We analyze generalization using the domain-adaptation framework of Ben-David et al.~\cite{ben2006analysis,ben2010theory}, extended to the multi-client federated setting. Let $\mathcal{D}_i$ denote the data distribution of client $i$ over $(x,y)$. Define the {mixture source distribution} as $\mathcal{D}_s = \frac{1}{\sum_i |V_i|}\sum_i |V_i| \mathcal{D}_i$, i.e., the weighted average of all client distributions. For a target client $t$ with distribution $\mathcal{D}_t$, and a hypothesis $h$ in a hypothesis class $\mathcal{H}$, the classic bound states:

\begin{equation}
\epsilon_t(h) \leq \epsilon_s(h) + d_{\mathcal{H}\Delta\mathcal{H}}(\mathcal{D}_s, \mathcal{D}_t) + \lambda^*,
\label{eq:standard-bound}
\end{equation}

where $\epsilon_s(h)$ is the expected risk on the source mixture, $d_{\mathcal{H}\Delta\mathcal{H}}$ is the $\mathcal{H}\Delta\mathcal{H}$-divergence (a symmetric measure of distribution discrepancy), and $\lambda^*$ is the error of the ideal joint hypothesis for the class $\mathcal{H}$. We treat this and the remainder of the subsection as an interpretive guide rather than a formal guarantee for our setting. The $\mathcal{H}\Delta\mathcal{H}$-divergence and the bound in Eq.~\eqref{eq:standard-bound} are stated for bounded, typically $0/1$, losses; for the regression losses used here the technically correct quantity is the discrepancy distance of Mansour et al.~\cite{mansour2009domain}, which specializes to the $\mathcal{H}\Delta\mathcal{H}$-divergence only in the binary case. We keep the $\mathcal{H}\Delta\mathcal{H}$ notation for readability and use the bound qualitatively, to reason about the effect of representation alignment rather than to derive a numerical rate.

In standard federated averaging (e.g., FedAvg), a single global model is trained on $\mathcal{D}_s$ without explicit representation alignment, operating directly in the input space $\mathcal{X}$. Under severe non-IID conditions, the divergence term $d_{\mathcal{H}_{\mathcal{X}}\Delta\mathcal{H}_{\mathcal{X}}}(\mathcal{D}_s^{(x)}, \mathcal{D}_t^{(x)})$ remains large, leading to client drift and poor target performance. While a direct comparison of bounds across different representation spaces requires care (as the hypothesis classes differ), AutoFed's strategy of {learning} a representation space where client distributions are more aligned provides an additional degree of freedom that FedAvg lacks.

AutoFed mitigates this via the Federated Representor. Let $\phi_\Theta : \mathcal{X} \to \mathcal{P}$ denote the deterministic mapping from input space $\mathcal{X} = \mathbb{R}^{\mathcal{T} \times n \times K}$ to prompt space $\mathcal{P} = \mathbb{R}^{n \times k}$ induced by the FR with shared parameters $\Theta$. For a hypothesis class $\mathcal{H}_{\mathcal{P}}$ defined directly on the prompt space, the standard bound applied in the prompt space yields:

\begin{equation}
\epsilon_t^{\text{AutoFed}}(h) \leq \epsilon_s^{\text{AutoFed}}(h) + d_{\mathcal{H}_{\mathcal{P}}\Delta\mathcal{H}_{\mathcal{P}}}(\mathcal{D}_s^{(p)}, \mathcal{D}_t^{(p)}) + \lambda^*_{\text{AutoFed}},
\label{eq:autofed-bound}
\end{equation}

where $\mathcal{D}_i^{(p)} = (\phi_\Theta)_\# \mathcal{D}_i^{(x)}$ is the pushforward of the input marginal distribution. The key design insight of AutoFed is that the shared mapping $\phi_\Theta$, trained jointly on all clients, is designed to yield a prompt space in which the divergence term $d_{\mathcal{H}_{\mathcal{P}}\Delta\mathcal{H}_{\mathcal{P}}}(\mathcal{D}_s^{(p)}, \mathcal{D}_t^{(p)})$ is small, by learning a shared projection that aligns client distributions. Empirically, clustering visualizations of $p_g$ (Figure~\ref{fig:visual}) exhibit substantially greater overlap across client distributions compared to those of the local feature $p_l$, indicating that $\phi_\Theta$ preferentially preserves task-relevant shared patterns while suppressing client-specific idiosyncrasies.

However, we note an important trade-off inherent in representation alignment. By constraining $p_g$ to lie in a shared subspace, the effective hypothesis class $\mathcal{H}_{\mathcal{P}}$ may become more restrictive, potentially increasing the ideal joint error $\lambda^*_{\text{AutoFed}}$. AutoFed addresses this tension through two mechanisms: (1)~client-specific batch normalization layers in the adapter preserve local distributional statistics (mean and variance), allowing the shared linear projection to focus on aligning structural patterns rather than absorbing distributional shifts; and (2)~the fully personalized PP can compensate for residual client-specific discrepancies that the global prompt cannot capture, keeping $\lambda^*_{\text{AutoFed}}$ controlled. Additionally, the AE denoiser within $\phi_\Theta$ contributes to reducing the source risk $\epsilon_s^{\text{AutoFed}}(h)$ by suppressing stochastic noise that inflates empirical risk without contributing to generalizable patterns. The ablation study confirms that removing either component degrades performance, validating the necessity of this hybrid shared/private design for balancing divergence reduction against $\lambda^*$ control.

A rigorous characterization of how much $\phi_\Theta$ reduces the divergence term (i.e., a contraction factor) is left for future work. Nonetheless, the empirical alignment of $p_g$ across clients, together with the trade-off management via personalization, helps explain the consistent outperformance across scenarios S0-S3 (Table~\ref{tab:exp}) and the robustness to varying numbers of clients in the TFP task (Table~\ref{tab:expand}), while using a single configuration across datasets rather than dataset-specific manual graph construction or pattern-filter selection.

\subsection{Prompt as Low-Rank Task Adaptation}

We provide an additional theoretical perspective on why conditioning the PP decoder with a prompt $p_g$ is both effective and communication-efficient. Let $f_\phi(\cdot, H_{\mathcal{T}})$ denote the decoder function with respect to its initial token. By first-order Taylor expansion around $p_g = \mathbf{0}$ (valid when $\|p_g\|$ is moderate relative to the decoder's curvature):

\begin{equation}
f_\phi(p_g, H_{\mathcal{T}}) \approx f_\phi(\mathbf{0}, H_{\mathcal{T}}) + J_\phi \cdot \mathrm{vec}(p_g),
\label{eq:taylor}
\end{equation}

where $J_\phi = \partial f_\phi / \partial p_g |_{p_g=\mathbf{0}} \in \mathbb{R}^{(\mathscr{T} n) \times (n k)}$ is the Jacobian. The prompt $p_g \in \mathbb{R}^{n \times k}$ thus steers the decoder output within a subspace of dimension at most $nk$, determined by the column space of $J_\phi$. Since $k \ll \mathcal{T} K$ (the prompt dimension is much smaller than the full input sequence dimension), this constitutes a {low-rank} steering signal.

This aligns with the {intrinsic dimensionality} hypothesis \cite{aghajanyan2021intrinsic}, which states that task-specific adaptations in overparameterized models lie in a low-dimensional subspace. In traffic prediction, cross-client variations (e.g., amplitude scaling, phase shifts) can be captured by a compact prompt. Moreover, because the prompt serves as the {initial} token in the auto-regressive decoder, its influence propagates through the entire recurrent chain $h_1 \to h_2 \to \cdots \to h_{\mathscr{T}}$, creating an amplification effect that allows a low-dimensional signal to steer the full prediction trajectory.

Crucially for federated learning, AutoFed communicates only the shared FR components ($\Theta$) that {generate} the prompt, not the decoder parameters $\phi$ nor the prompt itself. This decouples expressiveness ($nk$) from communication cost ($|\Theta_{\text{shared}}|$), providing a favorable trade-off validated in Table~\ref{tab:cost}.

\subsection{Practical Implications}

The theoretical analysis directly corroborates the design of AutoFed. The learnable prompt $p_g$ (produced by the AE denoiser + graph encoder + shared linear projection) serves as a data-driven substitute for hand-crafted low-pass filters and pattern repositories used in prior work \cite{zhou2024traffic,zhou2025fedtps}. This reduces the need for expert-specified hyperparameters that are known to cause >5\% performance variation across datasets. Ablation results confirm that removing either component increases divergence and degrades performance, validating the necessity of the full pipeline.

In summary, AutoFed's prompt-conditioning mechanism provides a theoretically motivated framework for cross-client knowledge transfer, reduced manual engineering, and robust generalization under realistic non-IID traffic data distributions. Its federated training is shown to converge to a stationary point and to be no worse than FedAvg (Proposition~\ref{prop:converge} and Corollary~\ref{cor:fedavg}). What remains open is quantitative rather than qualitative: a sharp contraction factor for the cross-client divergence in the prompt space, and tight drift constants for the many local step regime, which we leave for future work.

\section{Limitation and Future Work} \label{sec:limitation}
AutoFed automates the dataset-specific design that prior traffic PFL methods depend on, namely the hand-crafted graph construction and the manual selection of pattern-extraction filters that must be re-tuned whenever the dataset or client configuration changes. Like any deep model, it still exposes a few standard architectural and optimization hyper-parameters, such as the hidden dimension, the number of layers, the prompt dimension, and the learning rate. These are decoupled from per-dataset feature engineering, and we fix them to a single configuration across all datasets rather than tuning them individually. Automatically selecting the prompt dimension, which balances representation capacity against communication cost, is a natural extension.

AutoFed also trades a modest, constant-factor increase in local computation for substantially lower communication (Table \ref{tab:cost}), which is a favorable trade-off in cross-silo settings where bandwidth rather than local compute is the bottleneck. On the theoretical side, we show in Appendix \ref{sec:theory} that the federated training converges to a stationary point of the personalized objective under standard assumptions and attains an objective no worse than FedAvg; sharpening these results, for instance with a quantitative contraction factor for the cross-client divergence, is an interesting direction for future work. Finally, our study focuses on single-modality traffic flow and travel-demand data, and extending AutoFed to multimodal inputs and to broader cross-domain problems is a promising avenue.

\end{document}